\documentclass[10pt,twocolumn,letterpaper]{article}

\usepackage{iccv}
\usepackage{graphicx}
\usepackage{amsmath}
\usepackage{amssymb}
\usepackage{booktabs}
\usepackage{times}
\usepackage{epsfig}
\usepackage{array}
\usepackage{pifont}
\usepackage{multirow}
\usepackage{cite}

\usepackage[ruled]{algorithm2e}
\usepackage[table]{xcolor}
\usepackage[accsupp]{axessibility}

\usepackage{pdfpages}
\usepackage{multido}

\usepackage[breaklinks,colorlinks]{hyperref}
\usepackage[capitalize]{cleveref}

\newcommand\blfootnote[1]{%
  \begingroup
  \renewcommand\thefootnote{}\footnote{#1}%
  \addtocounter{footnote}{-1}%
  \endgroup
}

\crefname{section}{Sec.}{Secs.}
\Crefname{section}{Section}{Sections}
\Crefname{table}{Table}{Tables}
\crefname{table}{Tab.}{Tabs.}
\iccvfinalcopy 

\newcommand{\notsosmall}{\fontsize{10.5pt}{12pt}\selectfont}

\def\iccvPaperID{5450} 

\definecolor{somegray}{rgb}{0.5, 0.5, 0.5}
\newcommand{\darkgrayed}[1]{\textcolor{somegray}{#1}}
\makeatletter
\newcommand*\titleheader[1]{\gdef\@titleheader{#1}}
\AtBeginDocument{%
  \let\st@red@title\@title
  \def\@title{%
    \vskip-3em
    \bgroup\normalfont\large\centering\@titleheader\par\egroup
    \vskip1.5em\st@red@title}
}
\makeatother

\titleheader{\darkgrayed{This paper has been accepted for publication at the \\
IEEE/CVF International Conference on Computer Vision (ICCV), Paris, 2023.
\copyright IEEE}}

\title{GasMono: Geometry-Aided Self-Supervised Monocular Depth \\
Estimation for Indoor Scenes}

\begin{document}

\author{$^1$Chaoqiang Zhao \hspace{1.2cm} $^2$Matteo Poggi \hspace{1.2cm} $^2$Fabio Tosi \\ $^1$Lei Zhou \hspace{1.2cm} $^1$Qiyu Sun \hspace{1.2cm} $^{1,*}$Yang Tang \hspace{1.2cm} $^2$Stefano Mattoccia \vspace{0.3cm}\\
\notsosmall $^1$East China University of Science and Technology \hspace{1cm} $^2$University of Bologna 
}

\maketitle

\begin{abstract}
This paper tackles the challenges of self-supervised monocular depth estimation in indoor scenes caused by large rotation between frames and low texture. We ease the learning process by obtaining coarse camera poses from monocular sequences through multi-view geometry to deal with the former. However, we found that limited by the scale ambiguity across different scenes in the training dataset, a na\"ive introduction of geometric coarse poses cannot play a positive role in performance improvement, which is counter-intuitive.
To address this problem, we propose to refine those poses during training through rotation and translation/scale optimization.
To soften the effect of the low texture, we combine the global reasoning of vision transformers with an overfitting-aware, iterative self-distillation mechanism, providing more accurate depth guidance coming from the network itself.
Experiments on NYUv2, ScanNet, 7scenes, and KITTI datasets support the effectiveness of each component in our framework, which sets a new state-of-the-art for indoor self-supervised monocular depth estimation, as well as outstanding generalization ability. Code and models are available at \url{https://github.com/zxcqlf/GasMono}
\end{abstract}

\section{Introduction}
\blfootnote{$^*$ Corresponding author, yangtang@ecust.edu.cn.}
Depth estimation from images is one of the fundamental tasks in computer vision and plays a key role in several higher-level applications \cite{tang2022perception,liu2022accurate}. It has a long history and has been intensively studied building upon multi-view geometry \cite{hartley2003multiple,qiao2022improving}, exploiting image matching across two or multiple images and their camera positions \cite{poggi2021synergies,colmap,orbslam}.
The advent of deep learning rejuvenated this field and introduced new, exciting perspectives. Among many, the possibility of learning to estimate depth out of a single image -- for long considered the holy grail in computer vision -- became true \cite{eigen2014depth,laina2016deeper,fu2018deep,Ranftl2020,Ranftl2021}. However, this came at the cost of requiring a massive amount of images annotated with ground truth depth, often expensive to collect.

\begin{figure}[t]
\begin{center}
 \includegraphics[width=\linewidth]{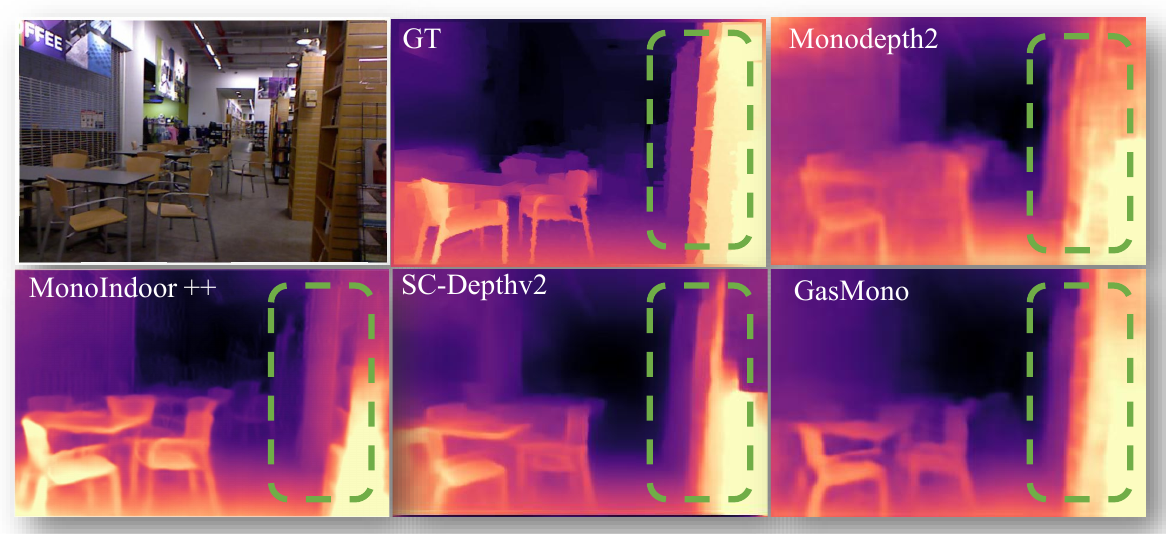}
\end{center}
   \caption{\textbf{Comparison between existing methods \cite{monodepth2,scdepthv2,monoindoor++} and GasMono.} Our framework shows remarkable accuracy on thin objects and global structures.}
\label{fig:fig1}
\end{figure}

In light of this, multi-view geometry maintained a key role in softening this latter constraint, allowing for the development of self-supervised monocular depth estimation frameworks \cite{zhao2020monocular}. These replace the need for depth labels by exploiting a proxy signal based on image reconstruction of the frame given as input to the network, i.e. the \textit{target}, starting from one or multiple \textit{source} images. The only requirement consists of collecting raw video sequences \cite{zhou2017unsupervised} or rectified stereo images \cite{godard2017unsupervised}. Among the two alternatives, the former \cite{zhou2017unsupervised} results as the cheapest and most flexible, since requiring a single camera only to move around and collect training data, with the relative camera poses between consecutive frames needing to be estimated by employing a dedicated pose network (PoseNet) alongside the depth estimation task. A large body of literature built ever more accurate self-supervised solutions, mainly focusing on outdoor environments -- i.e. with driving context \cite{kitti,drivingstereo} representing the preferred benchmark.

Nonetheless, the indoor setting is equally important for the development of, among others, navigation and assistive technologies, although featuring 1) much more complex ego-motion configurations and 2) large untextured regions, making this setting itself challenging for self-supervised depth estimation frameworks \cite{zhou2019moving}. 
Specifically, the different data collection equipment involved in outdoor and indoor scenes -- car-mounted vs handheld cameras -- leads to motion models largely different in the two cases. As an example, the average rotation between consecutive images in KITTI \cite{kitti} is $0.25^{\circ}$, while on NYUv2 \cite{nyuv2} dataset it is $2.28^{\circ}$ \cite{scdepthv2}.
Larger rotations among the images hinder the training process, because of the discontinuous Euler angle representation \cite{zhou2019continuity} commonly used for this task.
Additionally, the lack of texture tampers the image reconstruction process through which supervision is provided to the network, with several local minima in the training loss signal.

In this paper, we propose a novel, \textbf{G}eometry-\textbf{a}ided \textbf{s}elf-supervised framework for \textbf{Mono}cular depth estimation, dubbed \textit{GasMono}, specifically designed to face these challenges. Specifically, we leverage classic structure-from-motion algorithms such as COLMAP \cite{colmap} on the training sequences to initialize the pose estimation process, which is then refined to cope with the scale inconsistency occurring between the different monocular sequences part of the training set. 
To deal with the reduced texture characterizing indoor images, we combine recent architectures based on vision transformers \cite{monovit} with an iterative, self-distillation scheme to obtain stronger supervision and, thus, train GasMono more effectively. 
Our main contributions are:

\begin{itemize}
\item We tackle the challenges of learning camera poses in indoor scenes by exploiting strong priors coming from classic structure-from-motion algorithms \cite{colmap}.
\item This, however, is not sufficient: we explore the factors in such an approach making the training process unstable and a further, learning-based refinement strategy is proposed to optimize both rotation and translation/scale of the initial poses.
\item We explore the effectiveness of transformer architecture in improving the depth estimation of low-texture regions for indoor scenes, coupled with an overfitting-aware iterative self-distillation method, iteratively distilling pseudo labels from the depth network itself.
\item Our GasMono framework is evaluated on a variety of indoor datasets, establishing a new state-of-the-art for indoor, self-supervised monocular depth estimation. 
\end{itemize}
Fig. \ref{fig:fig1} shows a comparison between existing frameworks and GasMono, which shows more coherent predictions.

\section{Related Work}
We review the basic framework and related works concerning self-supervised monocular depth estimation.  

\textbf{Outdoor Self-Supervised Methods.}
Pivotal works in self-supervised monocular depth estimation focused on the outdoor setting, in particular dealing with driving environments \cite{kitti}, by replacing ground truth labels with image reconstruction losses on stereo images \cite{godard2017unsupervised} or monocular videos \cite{zhou2017unsupervised}. A rich literature of method following either the former \cite{3net18,Tosi_2019_CVPR,watson2019self,gonzalezbello2020forget,gonzalez2021plade,zhang2023qa} or the latter \cite{yin2018geonet,guizilini20203d,monodepth2,bian2021unsupervised,monovit} strategies have been developed, with monocular sequences being more practical, yet requiring 1) to estimate relative poses between frames at training time and 2) to deal with independently moving objects, violating the principles behind image reconstruction. 
This second challenge has been the object of study by several works, by masking the reprojection loss \cite{sun2021unsupervised,monodepth2}, exploiting additional tasks -- e.g. as optical flow \cite{yin2018geonet}, semantic segmentation \cite{klingner2020self} or both \cite{tosi2020distilled} -- while additional constraints such as the consistency between predicted depth maps \cite{bian2021unsupervised} or normals \cite{Yang_CVPR_2018}, adversarial losses \cite{zhao2020masked} and uncertainty modeling \cite{poggi2020uncertainty} result in good cues to ease the learning process.
Prior work \cite{Klodt_2018_ECCV} already shows how Structure-from-Motion can improve the learning process on videos, yet we will show how it cannot be exploited seamlessly in the indoor setting.
On an orthogonal dimension, the development of new architectures also plays a crucial role in boosting the accuracy of self-supervised depth estimation. Among them, PackNet \cite{guizilini20203d} and HRDepth \cite{hrdepth} proved consistent improvements over UNet-like architectures using vanilla encoders such as VGG and ResNet.
More recently, vision transformers \cite{monovit} yielded further improvements thanks to joint local and global reasoning, allowing for predicting more detailed depth maps on outdoor scenes compared to established CNNs.

\textbf{Indoor Self-Supervised Methods.}
Unlike the outdoor setting considered by early approaches, monocular sequences collected in indoor environments expose much more complex motions between frames, with much larger rotational components.
Because of Euler angle representation of 3D rotation hampering the learning process of the PoseNet \cite{zhou2019continuity}, the self-supervised framework being effective in outdoor environments cannot achieve satisfactory results on indoor scenes. To soften this problem, Zhao~\etal \cite{zhao2020towards} recover relative pose from dense optical flow to fully replace the PoseNet. Bian~\etal \cite{bian2021unsupervised,scdepthv2} demonstrate the role of rotation in unsupervised training, and they propose an Auto-Rectify Network to estimate and eliminate the rotation in advance. MonoIndoor series~\cite{ji2021monoindoor,monoindoor++} consider the depth range changes between different indoor scenes and design a transformer for depth factorization. Moreover, a residual pose estimation module is proposed in~\cite{ji2021monoindoor,monoindoor++} to iteratively optimize the relative pose predicted by networks. 
A second challenge when dealing with indoor scenes consist of the much lower texture present in images, making the image reconstruction loss often ineffective. To tackle this, Li~\etal \cite{li2021structdepth} use Manhattan normal constraint, while, Zhou~\etal \cite{zhou2019moving} propose to use optical flow to both obtain stronger supervision in low-textured regions, as well as to assist the training of the PoseNet. 

Our framework faces both challenges, respectively 1) by exploiting geometry by means of structure-from-motion algorithms \cite{colmap} to ease the pose estimation process and 2) by deploying vision transformers. However, these two strategies alone are ineffective: although being a good initialization, coarse poses \cite{colmap} are noisy and scale-inconsistent across different sequences. Thus, they need to be properly refined and further optimized. 

\textbf{Self-Distillation.}
Different from the knowledge distillation methods, self-distillation approaches treat the model as its own teacher and distill labels by itself. This strategy is widely used in the classification task \cite{ji2021refine} and, more recently, has been introduced for self-supervised depth estimation as well \cite{peng2021excavating} by exploiting stereo images during training. 
Inspired by this latter work, we introduce self-distillation into the monocular sequence-based self-supervised framework and prove that this approach alone yields sub-optimal improvements. 
Indeed, we cast self-distillation as an iterative approach by generating and selecting more and more accurate pseudo-labels. 

\section{GasMono Framework}
\begin{figure*}[t]
\begin{center}
 \includegraphics[width=\linewidth]{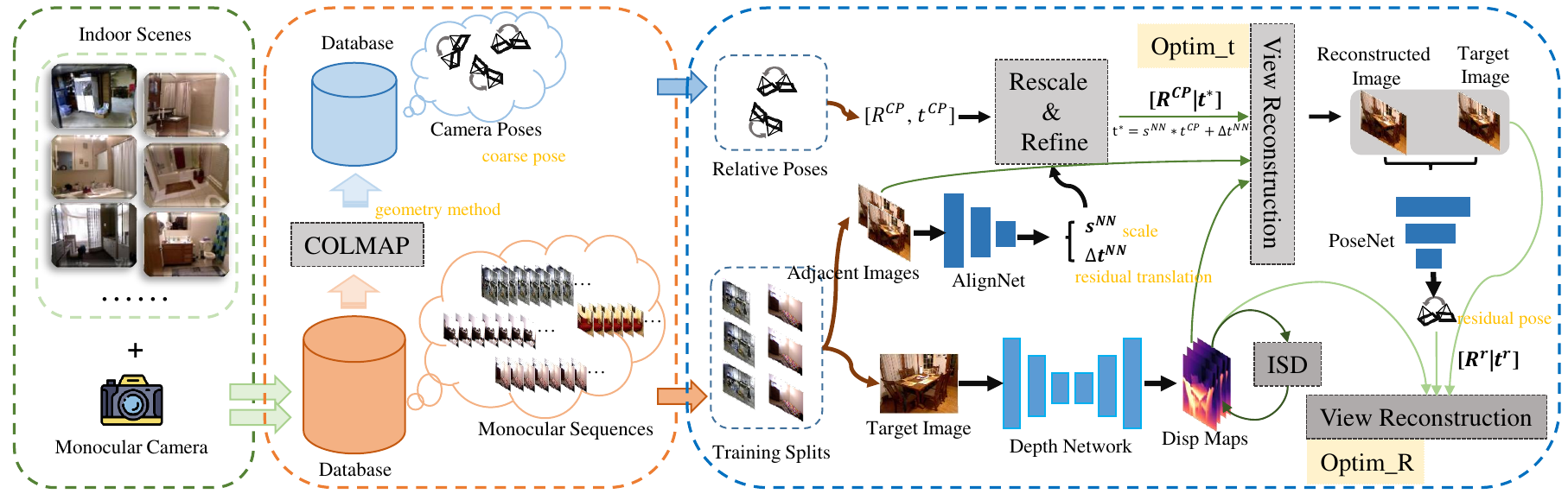}
\end{center}
   \caption{
   \textbf{GasMono: a geometry-aided self-supervised monocular depth estimation framework for indoor scenes.} Note that no ground truth labels are used in the training process. With the monocular image sequences selected by the camera from several indoor scenes, the structure-from-motion package COLMAP is used to estimate the coarse poses for the camera on each sequence. Then, the image sequences and coarse poses are used to train the depth model. To improve the coarse translation, an AlignNet is designed to estimate the scale $s^{NN}$ and residual translation $\Delta t$. Moreover, a PoseNet is also designed to further improve the pose, especially the coarse rotation, based on the reconstructed and target images. AlignNet and PoseNet are only used during training. }
\label{fig:frm}
\end{figure*}

We now introduce the key component of our framework, sketched in Fig. \ref{fig:frm}. 

\subsection{Geometry-Aided Pose Estimation}
The standard supervision protocol for self-supervised monocular depth estimation frameworks trained on video sequences consists of reprojecting pixels from a source image $\mathcal{I}_s$ into the target $\mathcal{I}_t$ according to estimated depth $D_t$ and relative camera pose $E_{t \rightarrow s} = R_{t \rightarrow s} | t_{t \rightarrow s}$. This means that, for pixel $p_t$ in the target view, its coordinates $p_s$ in the source view can be obtained as follows:
\begin{equation}
p_{s} \sim K [R_{t \rightarrow s}D_{t}(p_{t})K^{-1}p_{t}+t_{t \rightarrow s}] \label{eq:proj}
\end{equation}

Given the challenges of learning an accurate relative pose between images in presence of large rotations, we propose to get rid of the commonly used PoseNet and replace it with a traditional pose estimation algorithm. For this purpose, we leverage COLMAP \cite{colmap} to obtain camera poses $E_i^{CP}= R_i^{CP}|t_i^{CP}$ for images $\mathcal{I}_i$ in every single indoor sequence in the training set. Then, for a given image pair $\mathcal{I}_t$, $\mathcal{I}_{s}$, respectively the target and source frames, we can obtain the relative pose between the two as $E_{t\rightarrow s}^{CP} = R_{t\rightarrow s}^{CP}|t_{t\rightarrow s}^{CP} = E_{s}^{CP} {E^{CP}_t}^{-1}$. 
Differently from two-frame pose estimation, a structure-from-motion pipeline such as COLMAP reasons globally over the entire sequence.
We argue that, since pose estimation represents a side task for learning monocular depth, it is worth exploiting sequences as a whole.

Nonetheless, poses estimated by COLMAP, which we will refer to as \textit{coarse} from now on, suffers from some problems, specifically 1) scale inconsistency across different sequences and scale drift within the very same scene because of monocular ambiguity, 2) noise in estimated rotation and translation due to the lack of texture. This makes COLMAP alone ineffective at seamlessly replacing the PoseNet to train a monocular depth network.

\subsubsection{Translation Rescaling and Refinement}
To solve the former issue, 
we deploy a shallow network, AlignNet, to refine the coarse translation and to rescale it during training, allowing to overcome scale inconsistency across the many sequences in the training set. 

Accordingly, AlignNet processes the target and source images and predicts a scale factor $s^{NN}$ and a residual shift $\Delta t^{NN}$ applied to the translation component $t_{t \rightarrow s}^{CP}$ estimated by COLMAP. Then, estimated translation $t_{t\rightarrow s}$ from target to source views is obtained as 
\begin{equation}
t^*_{t\rightarrow s} = s^{NN}_{t \rightarrow s}t_{t \rightarrow s}^{CP} + \Delta t^{NN}_{t \rightarrow s} \label{eq:rescalet}
\end{equation}
This refined translation vector is then used in Eq. \ref{eq:proj}, leading to scale adjustment when learning to estimate monocular depth maps by exploiting $R_{t \rightarrow s}^{CP}|t^*_{t \rightarrow s}$, both within the same sequence as well as across all the scenes in the training dataset, shown as ``Optim\_t'' in Fig. \ref{fig:frm}. 

We can consider AlignNet as a training optimization tool, refining coarse poses to make them scale-consistent as a whole. As such, it loses its utility once the training procedure is completed.

\subsubsection{Rotation Optimization}
Bian~\etal \cite{scdepthv2} has proved that the rotation in self-supervised training is independent of depth learning. Nonetheless, in case of inaccurate rotations being estimated between target and source frames, noisy gradients will arise and affect the training of the depth network as well.
However, we argue that such an error is a small residual rotation, largely smaller compared to the original rotation between the two frames, and thus a deep network could learn such a correction.

Inspired by the rotation and pose rectify methods in recent works \cite{scdepthv2,ji2021monoindoor}, we re-introduce the PoseNet to solely estimate the residual pose $R_{t \rightarrow s}^r | t_{t \rightarrow s}^r$ between the image being reconstructed from $R_{t \rightarrow s}^{CP}| t^*_{t \rightarrow s}$ and the target one, by computing an additional reprojection loss using the same, estimated depth and the newly computed, residual pose -- ``Optim\_R'' in Fig. \ref{fig:frm}. 
Although the PoseNet cannot deal properly with large rotations, it can effectively estimate the small, residual rotations result of the COLMAP errors.

\subsection{Low-Texture Regions}
In self-supervised training, back-propagation behavior replies to the photometric gradient changes on RGB images. Those regions with effective photometric changes provide strong gradients for depth learning, while regions with low-texture regions, like walls and floors, cannot provide effective supervisory signals, because multiple depth hypotheses lead to photometric errors being close to zero~\cite{zhou2019moving}, thus driving the network into local minima. Therefore, the depth estimation process mainly relies on the inferring ability of the network itself for low-texture regions. The use of some additional constraints, coming from cues such as optical flow or plane normals  \cite{li2021structdepth,zhou2019moving}, might help. Nonetheless, 
this requires additional supervision and can suffer the same issues due to low texture, as in the case of optical flow.
Thus, we choose to tackle it on the architectural side, in particular by overcoming the limited receptive fields of CNNs with a vision transformer. 
Moreover, previous works proved the effectiveness of label distillation \cite{poggi2020uncertainty,peng2021excavating} to improve the accuracy of the depth network. Therefore, we propose a more effective solution for obtaining more accurate distilled labels in order to provide stronger supervision to GasMono.

\subsubsection{Network Architecture}
Our framework consists of three networks, a Depth Network for monocular depth estimation, an Alignment Network (AlignNet) for scale correction and residual translation prediction, and a PoseNet for residual pose estimation. The overall training architecture is shown in Fig. \ref{fig:frm}.

Considering the outstanding performance of the transformer in long-range
relationship modeling between features, in order to enhance the global feature extraction for the depth inferring of low-texture regions, we introduce a transformer encoder, MPViT \cite{mpvit}, as the depth encoder by following \cite{monovit}. The self-attention mechanism in the encoder is implemented in an efficient factorized way~\cite{mpvit}:
\begin{equation}
    \text{FactorAtt}(\mathbf{Q}, \mathbf{K}, \mathbf{V}) = \frac{\mathbf{Q}}{\sqrt{C}}(\text{softmax}(\mathbf{K})^{T}\mathbf{V})
\end{equation}
where $C$ refers to the embedding dimension. Query ($\mathbf{Q}$), key ($\mathbf{K}$), and value ($\mathbf{V}\in \mathbb{R}^{N\times C}$) vectors are projected from vision tokens.
Moreover, for the depth decoder, we replace the standard upsampling implemented by Monodepth2 \cite{monodepth2} and others with convex upsampling \cite{raft}, bringing the 4 scale disparity maps to full resolution, at which they are used for the iterative self-distillation operations described in the remainder.
To build both the AlignNet and PoseNet, we use the same architecture used to implement the PoseNet in previous works \cite{monodepth2,monovit}, which is based on ResNet-18 \cite{resnet}.

\subsubsection{Iterative Self-Distillation Method}
Inspired by \cite{peng2021excavating}, we propose an overfitting-driven Iterative Self-Distillation procedure (ISD) to obtain the depth map with the lowest per-pixel minimum reprojection error, yielding more accurate labels for any specific training sample. 

The key steps of ISD are listed in Algorithm \ref{alg:isd}.
For each single training image, we iterate this process multiple times (row 4). At the first iteration, we select the per-pixel minimum reprojection error across all scales and its corresponding predicted depth (rows 6-14). Then, we update the network by minimizing a depth loss between the current best depth map and predictions at each scale (rows 15-16). This procedure is repeated for a number of iterations.

We will show in our experiments how this allows for improving the supervision provided by na\"ive distillation  \cite{peng2021excavating}.

\begin{algorithm}[t]  
    
	\caption{Iterative Self-Distillation \textbf{ISD}}
\label{alg:isd}
	\LinesNumbered \KwIn{\\
 \textit{Target image:} $ \mathcal{I} \in R^{b\times3\times H \times W}$\; 
 \textit{Reconstructed target image:} $ \tilde{\mathcal{I}}^{s} \in R^{b\times3\times H\times W}$\; 
 \textit{Disparity map:} $ D^{s} \in R^{b\times1\times H\times W}$\; 
 }
	\KwOut{
 Self-distillation loss ($L_{ISD}$)}
    \textit{total\_iterations =} $n>0$\;
    \textit{curr\_iterations =} 0\;
    \textit{Dict =} \{\}\;
	\For{curr\_iteration $<$ total\_iterations}{
        \textit{curr\_iterations +=} 1\;
    	\For{s in range(4)}{
     Rec\_error = $ \mathcal{F}(\tilde{\mathcal{I}^{s}}, \mathcal{I})$ $ \in R^{b\times1\times H\times W}$\;
    		\If{Dict is empty}{
    			Dict[`disp\_best'] = $ D^{0}$\;
                Dict[`error\_min'] = Rec\_error\;}
                
           Dict[`disp\_best'] = torch.where( Rec\_error$<$ Dict[`error\_min'], $D^{s}$, Dict[`disp\_best'])\;
           Dict[`error\_min'] = torch.min( torch.cat( [Rec\_error, Dict[`error\_min']] ))\;
	}
      \scriptsize $L_{ISD}$ =  $\sum_{s=1}^{4} (log(|Dict[`disp\_best'] - D^{s}|+1))$\;
     \footnotesize $L_{tot}$.backward()
	}
\end{algorithm}

\subsection{Training Losses}
Different from competitors \cite{monoindoor++,scdepthv2}, we follow outdoor methods \cite{monodepth2}, the key term of our training loss consists of the minimum view reconstruction loss, accompanied by a smoothness term.

\textbf{View Reconstruction Loss.} 
Having obtained a reconstructed image $\tilde{\mathcal{I}}$, its error with respect to the target image $\mathcal{I}$ is measured by means of Structured Similarity Index Measure (SSIM) and L1 difference, combined as in~\cite{monodepth2}:

\begin{equation}
    \mathcal{F}(\tilde{\mathcal{I}}, \mathcal{I}) = \alpha \cdot \frac{1- \text{SSIM}(\tilde{\mathcal{I}}, \mathcal{I})}{2} + (1-\alpha) \cdot|\tilde{\mathcal{I}}-\mathcal{I}|
\end{equation}
with $\alpha$ commonly set to 0.85~\cite{monodepth2}. Besides, to soften the effect of occlusions between two views, the minimum among losses obtained with respect to forward and backward adjacent frames being warped is computed:

\begin{equation}
    \mathcal{L}_{rec}(p) = \min_{i \in [1,-1]} \mathcal{F}(\tilde{\mathcal{I}}_i(p), \mathcal{I}(p))
\end{equation}
with `1' and `-1' referring to the forward and backward adjacent frames, respectively.

\textbf{Smoothness Loss.}
The edge-aware smoothness loss is used to further improve the inverse depth map $d$:
\begin{equation}
\mathcal{L}_{smooth} = |\partial_{x} d^{*}|e^{\partial_{x} I} + |\partial_{y} d^{*}|e^{\partial_{y} I}, \label{eq:smooth}
\end{equation}
where $d^{*}=d/\hat{d}$ represents the mean-normalized inverse depth. Besides, following~\cite{monodepth2}, an auto-mask $\mu$ is calculated to filter static frames and some repeated texture regions.

\textbf{Iterative Self-Distillation Loss.}
As described before, GasMono self-distills proxy labels to be used as additional supervision. Given proxy labels $d^{best}$ obtained according to Algo. \ref{alg:isd}, we minimize the log-error of predicted depth $d$ with respect to it:
\begin{equation}
\mathcal{L}_{ISD} =log(|d^{best} - d|+1) \label{eq:dist2}
\end{equation}

\textbf{Total Loss.} Finally, the view reconstruction loss $\mathcal{L}_{rec}$, the smoothness loss $\mathcal{L}_{smooth}$ and the distillation term $\mathcal{L}_{ISD}$ are computed on outputs at any given scale -- brought to full resolution -- to obtain the total loss term $\mathcal{L}_{tot}$.
More specifically, two reconstruction losses are computed, respectively 
$\mathcal{L}^{Optim\_t}_{rec}$ and $\mathcal{L}^{Optim\_R}_{rec}$:
\begin{equation}
\begin{split}
\mathcal{L}_{tot} &= \mathcal{L}^{Optim\_t}_{rec} + \beta \cdot \mathcal{L}^{Optim\_R}_{rec} \\ &+ \lambda \cdot \mathcal{L}_{smooth} + \mu \cdot \mathcal{L}_{ISD}  \label{eq:total} 
\end{split}
\end{equation}
with $\mathcal{L}^{Optim\_t}_{rec}$ and $\mathcal{L}^{Optim\_R}_{rec}$ being computed with reconstructed images based on the poses obtained after translation and rotation optimizations respectively, and with $\beta, \lambda$ and $\mu$ being set to 0.2, 0.001 and 0.1.
Finally, total losses are averaged across scales.

\begin{table*}[ht]
\centering
\setlength
\tabcolsep{7pt}{
\scalebox{1}{
\footnotesize
\begin{tabular}{ll|c |c c c c | c c c}
\hline
&\multirow{2}{*}{Details} & \multirow{2}{*}{scale\_std} &\multicolumn{4}{c|}{lower is better} & \multicolumn{3}{c}{higher is better}\\
& & &Abs Rel$\downarrow$& Sq Rel$\downarrow$&  RMSE$\downarrow$ & RMSE log $\downarrow$& $\delta_{1}\uparrow$ & $\delta_{2} \uparrow$ & $\delta_{3} \uparrow$\\
\hline
(a) & $\textbf{T}(R^{NN}, t^{NN})$ -- Monodepth2 \cite{monodepth2}& 0.259 & 0.167 & 0.137 & 0.603 & 0.208 & 0.754 & 0.944 & 0.985\\
\hline
\hline
(b) & $\textbf{T}^{\ddagger}(R^{CP}, t^{CP})$ & 0.597 & 0.298 & 0.521 & 1.079 & 0.346 & 0.534 & 0.808 & 0.925\\
(c) & $\textbf{T}(R^{CP}, t^{NN})$ & 0.215 & 0.160 & 0.131 & 0.584 & 0.203 & 0.771 & 0.945 & 0.985\\
(d) & $\textbf{T}^{\ddagger}(R^{NN}, t^{CP})$ & - & 0.311 & 0.460 & 0.999 & 0.369 & 0.511 & 0.779 & 0.904\\
(e) & $\textbf{T}(R^{CP}, t^{CP}*s^{NN})$ & 0.237 & 0.183 & 0.169 & 0.661 & 0.231 & 0.718 & 0.920 & 0.978\\
(f) & $\textbf{T}^{\ddagger}(R^{CP}, t^{CP}+\Delta t^{NN})$ & 0.501 & 0.255 & 0.326 & 0.888 & 0.309 & 0.583 & 0.845 & 0.946\\
(g) & $\textbf{T}(R^{CP}, t^{CP}*s^{NN} + \Delta t^{NN})$ & 0.240 & 0.159 & 0.132 & 0.590 & 0.203 & 0.775 & 0.945 & 0.986\\
\hline
\hline
(h) &  Monodepth2 \cite{monodepth2}& * & 0.167 & 0.137 & 0.603 & 0.208 & 0.754 & 0.944 & 0.985\\
(i) &  + MPViT & * & 0.145 & 0.109 & 0.546 & 0.186 & 0.804 & 0.959 & 0.991\\
\hline
(j) &  + MPViT + CP & * & 0.311 & 0.460 & 0.999 & 0.369 & 0.511 & 0.779 & 0.904\\
(k) &  + MPViT + CP + Optim\_t& * & 0.124 & 0.090 & 0.490 & 0.165 & 0.850 & 0.968 & 0.991\\
(l) &  + MPViT + CP + Optim\_t + Optim\_R & * &   0.114  &   0.085  &   0.469  &   0.155  &   0.867  &   0.972  &   0.992\\
\hline
(m) &  + MPViT + ISD & * & 0.137 & 0.100 & 0.523 & 0.177 & 0.821 & 0.964 & 0.991\\
(n) &  + MPViT + CP + Optim\_t + ISD& * & 0.122 & 0.091 & 0.486 & 0.162 & 0.857 & 0.969 & 0.991\\
(o) &  + MPViT + CP + Optim\_t + Optim\_R + ISD& * &   0.113  &   0.083  &   0.459  &   0.153  &   0.871  &   0.973  &   0.992\\
\hline
\end{tabular}}}
\vspace{0.3cm}
\caption{ \textbf{Ablation Studies}. The upper part is based on the monodepth2 \cite{monodepth2}, which explores the utilization and optimization methods for the coarse pose from COLMAP. ``$(R/t)^{CP/NN}$'' refers to the $(R/t)$ generated by $COLMAP/Neural Network$. ``scale\_std'' stands for the std. error of the scale alignment factor on the testset. ``$^{\ddagger}$'' denotes that the training process has a high probability of not converging, and we only record the results of the converged case here. The other part is ablation experiments to prove the effectiveness of each module in this paper. $CP$ refers to the model trained by using coarse pose [$R^{CP},t^{CP}$].}
\label{tab:abl}
\end{table*}

\section{Experimental Results}

We now evaluate the performance of GasMono with respect to existing methods from the literature. We first describe the datasets and protocol used for our experiments, then conduct an exhaustive study on GasMono behavior when trained with COLMAP poses and our pose optimization strategies, followed by an ablation study on our model and a final, comparison with state-of-the-art approaches.

\subsection{Implementation Details}

We start by describing the datasets involved in our evaluation and the implementation details of our method.

\subsubsection{Datasets}

We conduct our experiments on three popular indoor datasets: NYUv2 \cite{nyuv2}, 7scenes~\cite{7scene}, and ScanNet~\cite{scannet}. To validate our iterative-self distillation strategy, we also evaluate on the outdoor KITTI \cite{kitti} dataset.

\textbf{Indoor Datasets:}
\textbf{NYUv2} contains 464 indoor sequences with dense depth ground truth captured by a handheld RGB-D camera at 640$\times$480 resolution. Following previous works~\cite{scdepthv2}, we use the officially provided 654 densely labeled images for testing and the remaining 335 sequences for training and validation. Since COLMAP failed in some short training sequences, these have been excluded. Differently from the MonoIndoor series \cite{ji2021monoindoor,monoindoor++}, which rectifies the image distortion with provided camera intrinsics, we directly use raw images for training.
\textbf{7-Scenes} consists of 7 indoor scenes, each containing several image sequences at 640$\times$480 resolution. We follow the official train/test split for each scene. Following \cite{scdepthv2}, we extract the first image from every 10 frames for each scene for testing.
\textbf{ScanNet} provides 1513 indoor RGB-D videos, captured by handheld devices. To evaluate the generalization performance of our depth model trained on NYUv2, we use the officially released test set following \cite{scdepthv2,monoindoor++}. On these datasets, images are resized to 320$\times$256 during both training and testing.

\textbf{Outdoor Dataset:}
\textbf{KITTI} contains 61 scenes, with a typical image size of $1242 \times 375$, captured using a stereo rig mounted on a moving car equipped with a LiDAR sensor. Following the literature~\cite{godard2017unsupervised,zhou2017unsupervised}, we use the image split of Eigen \textit{et al.}~\cite{Eigen2014} for training and validation. To compare with the existing solutions, we evaluate the model on the test split of \cite{Eigen2014} with the same evaluation metrics. Images are resized to 640$\times$192 during both training and testing.

\subsubsection{Settings}

We implement GasMono in PyTorch, training it for 40 epochs on the NYUv2 dataset and 20 epochs on the KITTI dataset by using AdamW~\cite{adamw} optimizer. The batch size is set to 12 and the number of iterations for ISD to 2.
The translation, pose encoder and depth encoders are pre-trained on ImageNet~\cite{imagenet}. 
For training, we use a single NVIDIA Tesla P40 GPU. Source code and generated coarse poses will be open-sourced in case of acceptance.

\subsection{Ablation Studies}

We start by inquiring about the impact of any component in GasMono, to assess their effectiveness at dealing with the challenges of indoor monocular depth estimation.

\begin{figure*}[t]
\begin{center}
\includegraphics[width=0.9\linewidth]{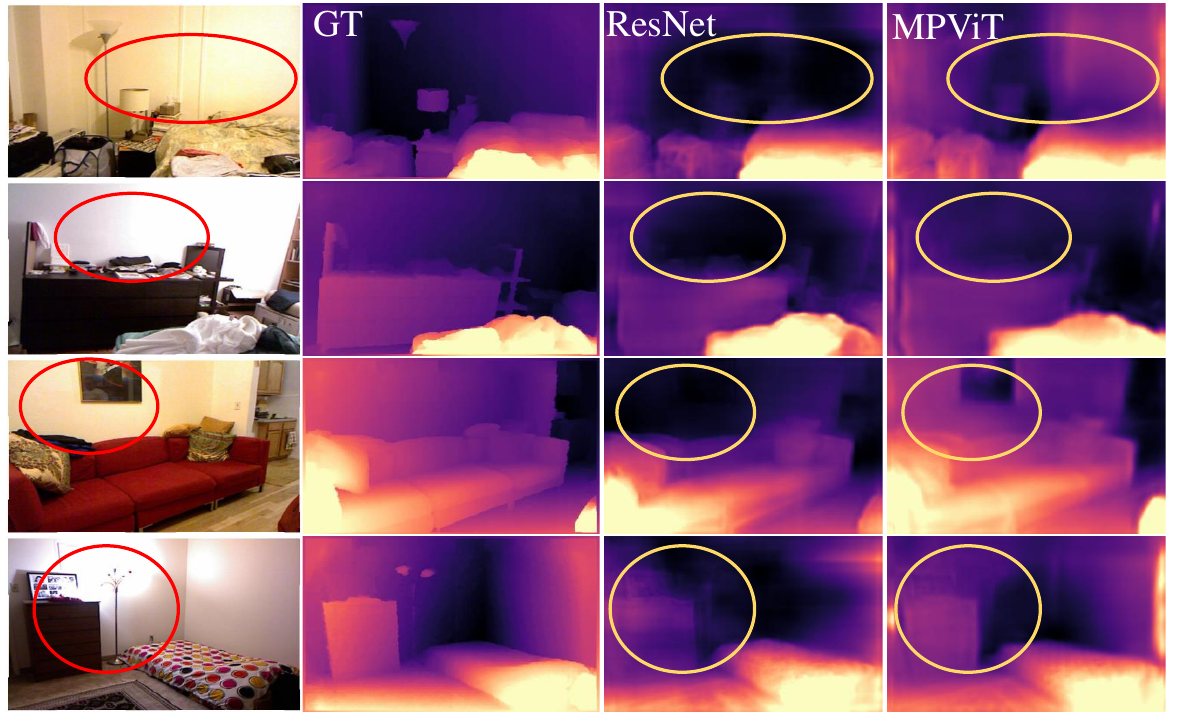}
\end{center}
   \caption{
   \textbf{Effects of different encoders on low-texture depth estimation.} 
   The transformer-based encoder, MPViT, can effectively improve the depth inferring on low-texture regions.}
\label{fig:abl}
\end{figure*}

\subsubsection{Coarse Poses and Network Convergence}

To analyze the difficulties as well as improvements yielded by employing geometry-based pose estimation algorithms such as COLMAP, we build a set of experiments based over the Monodepth2 baseline \cite{monodepth2} on the NYUv2 dataset~\cite{nyuv2}, collected in Tab. \ref{tab:abl}. For training, only photometric and smoothness losses from \cite{monodepth2} are used in any setting. 

Row (a) contains the results achieved by the baseline -- i.e., Monodepth2 using a PoseNet to estimate poses.
Then, rotation $R^{CP}$ and translation $t^{CP}$ estimated by COLMAP are directly used to replace the pose network in (b), showing that this approach does not yield satisfactory accuracy. 

Then, we select only one among rotation $R^{NN}$ and translation $t^{NN}$ components estimated by the PoseNet to respectively replace the COLMAP counterpart. From the results, we can notice that using only translation $t^{NN}$ estimated by the network, while keeping the rotation component being estimated by COLMAP (c), allows for outperforming the baseline. Besides, using the $R^{NN}$ and $t^{CP}$ (d) yields to the worst results observed so far. These two experiments confirm that 1) the translation component $t^{CP}$ is the main cause of failure when using COLMAP poses -- indeed, replacing it with $t^{NN}$ is sufficient to boost performance -- and 2) COLMAP estimated rotations results largely better than those learned through a PoseNet. 
Concerning $t^{CP}$, we ascribe two main factors preventing it from making the learning process work properly: scale and shift components.

\textbf{Scale Factor}: since the training set contains hundreds of monocular sequences, the scale ambiguity and scale drift caused by monocular SfM result in scale inconsistency of poses between different training samples, which affects the depth network during training as well. In the table, we report the standard deviation of the scale alignment factor on the test set (scale std), to reflect how depth scale is affected. We can notice how using COLMAP coarse poses (b) greatly increases the standard deviation, pointing out scale inconsistency across the scenes. By means of estimating a rescaling factor $s^{NN}$ (e), we can significantly reduce it, making it even lower with respect to the baseline and, consequently, improving the accuracy of the estimated depth map.

\textbf{Shift Factor}:
although poses are generated by COLMAP, they can be noisy and inaccurate. Therefore, we use a neural network to regress the residual translation $\Delta t^{NN}$ to be applied to $t^{CP}$.
Estimating this alone (f) results ineffective because of the scale ambiguity issue discussed so far. However, when both scale and shift are estimated (g), we achieve the best results in terms of depth accuracy.

\begin{table}[t]
\centering
\scalebox{0.75}{
\renewcommand{\tabcolsep}{2pt}
\begin{tabular}{l |c c c | c c c}
\hline
\multirow{2}{*}{Methods} &\multicolumn{3}{c|}{lower is better} & \multicolumn{3}{c}{higher is better}\\
  &Abs Rel$\downarrow$&  RMSE$\downarrow$ & RMSE log $\downarrow$& $\delta_{1} \uparrow$ & $\delta_{2} \uparrow$ & $\delta_{3} \uparrow$\\
\hline
MonoUncertainty\cite{poggi2020uncertainty}  & 0.111  & 4.756 & 0.188 & 0.881 & 0.961 & 0.982\\
HR-Depth~\cite{hrdepth}        & 0.109  & 4.632 & 0.185 & 0.884 & 0.962 & 0.983\\
CADepth~\cite{cadepth}   & 0.105 & 4.535 & 0.181 & 0.892 & 0.964 & 0.983\\
DIFFNet~\cite{diffnet}      & 0.102 & 4.445 & 0.179 & 0.897 & 0.965 & 0.983\\
DevNet~\cite{devnet}      & 0.100  & 4.412 & 0.174 & 0.893 & 0.966 & \textbf{0.985}\\
BRNet~\cite{brnet}  & 0.105  & 4.462 & 0.179 & 0.890 & 0.965 & 0.984\\
SC-Depthv2 \cite{scdepthv2} & 0.118 & 4.803 & 0.193 & 0.866 & 0.958  &0.981\\
\hline
Monodepth2 \cite{monodepth2}   & 0.115 & 4.863 & 0.193 & 0.877 & 0.959 & 0.981\\
 +  \textbf{SD}  &   0.112  &   4.814  &   0.191  &   0.879  &   0.860  &   0.981 \\ 
 +  \textbf{ISD}  &   0.111   &   4.732  &   0.189  &   0.880  &   0.961  &   0.982  \\  
\hline
MonoViT \cite{monovit} & 0.099 & 4.372 & 0.175 & 0.900 & 0.967 & 0.984\\
 + \textbf{SD} & 0.099 & 4.359 & 0.174 & 0.902 & 0.967 & 0.984\\
 + \textbf{ISD} & \textbf{0.098}  & \textbf{4.303} & \textbf{0.173} & \textbf{0.903} & \textbf{0.968} & 0.984\\
\hline
\end{tabular}}
\vspace{0.3cm}
\caption{\textbf{Testing our ISD with different baselines on the outdoor KITTI Dataset \cite{kitti}}; The depth network is trained by outdoor monocular image sequences in a self-supervised manner.}
\label{tab:kitti}
\end{table}

\begin{table}[t]
\centering

\scalebox{0.77}{
\renewcommand{\tabcolsep}{2pt}
\begin{tabular}{l |c| c c | c c c}
\hline
\multirow{2}{*}{Methods} & \multirow{2}{*}{Supervision}& \multicolumn{2}{c|}{Error Metric} & \multicolumn{3}{c}{Accuracy Metric}\\
  & &Abs Rel$\downarrow$&  RMSE$\downarrow$ & $\delta_{1} \uparrow$ & $\delta_{2} \uparrow$ & $\delta_{3} \uparrow$\\
\hline
      Make3D \cite{saxena2008make3d} & Sup. & 0.349 & 1.214 & 44.7 & 74.5 & 89.7 \\
      Li $\textit{et al.}$ \cite{li2017two} & Sup. & 0.143 & 0.635 & 78.8 & 95.8 & 99.1 \\
      Eigen $\textit{et al.}$ \cite{eigen2015predicting} & Sup. & 0.158 & 0.641 & 76.9 & 95.0 & 98.8\\
      Laina $\textit{et al.}$ \cite{laina2016deeper} & Sup. & 0.127 & 0.573 & 81.1 & 95.3 & 98.8 \\
      DORN \cite{fu2018deep} & Sup. & 0.115 & 0.509 & 82.8 & 86.5 & 99.2 \\
      AdaBins \cite{bhat2021adabins} & Sup. & \textbf{0.103} & 0.364 & 90.3 & 98.4 & 99.7\\
      DPT \cite{Ranftl2021} & Sup. & 0.110 & \textbf{0.357} & \textbf{90.4} & \textbf{98.8} & \textbf{99.8}\\
      \hline
      Zhou $\textit{et al.}$ \cite{zhou2019moving} & Self-Sup. & 0.208 & 0.712 & 67.4 & 90.0 & 96.8 \\
      Zhao $\textit{et al.}$ \cite{zhao2020towards} & Self-Sup. & 0.189 & 0.686 & 70.1 & 91.2 & 97.8 \\
      SC-Depthv1 \cite{bian2021unsupervised} & Self-Sup. & 0.157 & 0.593 & 78.0 & 94.0 & 98.4\\
      P$^2$Net+PP \cite{yu2020p} & Self-Sup. & 0.147 & 0.553 & 80.4 & 95.2 & 98.7 \\
      StructDepth \cite{li2021structdepth} & Self-Sup. & 0.142 & 0.540 & 81.3 & 95.4 & 98.8 \\
      MonoIndoor \cite{ji2021monoindoor} & Self-Sup. & 0.134 & 0.526 & 82.3 & 95.8 & 98.9 \\
      SC-Depthv2 \cite{scdepthv2} & Self-Sup. & 0.138 & 0.532 & 82.0 & 95.6 & 98.9\\
      MonoIndoor++ \cite{monoindoor++} & Self-Sup. & 0.132 & 0.517 & 83.4 & 96.1 & 99.0 \\
      DistDepth \cite{wu2022toward} & Self-Sup. & 0.130 & 0.517 & 83.2 & 96.3 & 99.0\\
      \textbf{GasMono (Ours)}  & Self-Sup. + Pose & \textbf{0.113} & \textbf{0.459} & \textbf{87.1} & \textbf{97.3} & \textbf{99.2} \\
      \hline

\end{tabular}}
\vspace{0.3cm}
\caption{\textbf{Evaluation on NYUv2~\cite{nyuv2}.} Sup.: trained with ground truth; Self-Sup.: trained on image sequences; Pose: uses COLMAP poses at training time.}
\label{tab:nyu}
\end{table}

\subsubsection{From Monodepth2 to GasMono}

Having assessed the impact of coarse poses and how to use them effectively, we now measure the improvements yielded by any single component differentiating GasMono from Monodepth2, by sequentially adding one component at a time. Results are collected in the bottom part of Tab. \ref{tab:abl}. 

\textbf{Transformer Encoder.} One of the main challenges in indoor environments concerns the lack of texture and the global receptive field of a transformer can help deal with it. By replacing the ResNet18 encoder in Monodepth2 (h) with a state-of-the-art vision transformer (i) -- MPViT \cite{mpvit} -- allows for much more accurate depth predictions.
Fig. \ref{fig:abl} shows some qualitative examples: the ResNet-based depth network cannot infer a reasonable disparity for large regions having low texture, whereas the transformer-based depth network benefits from the long-range feature and global feature modeling and yields much better predictions.
More ablation studies on the transformer encoder -- and the depth decoder -- are reported in the {{supplementary material}}.

\begin{table}[t]
\centering
\scalebox{0.7}{
\renewcommand{\tabcolsep}{2pt}
\begin{tabular}{l |c| c c | c c c}
\hline
    \multirow{2}{*}{Methods} &
    \multirow{2}{*}{Supervision} & \multicolumn{2}{c|}{Error Metric} & \multicolumn{3}{c}{Accuracy Metric} \\
    \cline{3-7}
  & &Abs Rel$\downarrow$&  RMSE$\downarrow$ & $\delta_{1} \uparrow$ & $\delta_{2} \uparrow$ & $\delta_{3} \uparrow$\\
    \hline
    Laina~\etal~\cite{laina2016deeper} & Sup. & 0.141 & 0.339 & 0.811 & .958 & 0.990 \\
    VNL~\cite{vnl} & Sup. & 0.123 & 0.306 & 0.848 & 0.964 & 0.991 \\
      DPT \cite{Ranftl2021} & Sup. & \textbf{0.089} & \textbf{0.220} & \textbf{0.917} & \textbf{0.985} & \textbf{0.997}\\
    \hline
    TrainFlow~\cite{zhao2020towards} & Self-Sup. & 0.179 & 0.415 & 0.726 & 0.927 & 0.980 \\
    SC-Depthv1~\cite{bian2021unsupervised} & Self-Sup. & 0.169 & 0.392 & 0.749 & 0.938 & 0.983 \\
    SC-Depthv2~\cite{scdepthv2} & Self-Sup. & 0.156 & 0.361 & 0.781 & 0.947 & 0.987 \\
    Monodepth2~\cite{monodepth2} & Self-Sup. & 0.170 & 0.401 & 0.730 & 0.948 & 0.991 \\
    MonoIndoor \cite{ji2021monoindoor} & Self-Sup. & 0.154 & 0.373 & 0.779 & 0.951 & 0.988 \\
    MonoIndoor++ \cite{monoindoor++} & Self-Sup. & 0.138 & 0.347 & 0.810 & 0.967 & \textbf{0.993} \\
    \textbf{GasMono (Ours)}  & Self-Sup. + Pose & \textbf{0.120} & \textbf{0.301} & \textbf{0.856} & \textbf{0.972} & \textbf{0.993} \\
    \hline

\end{tabular}}
\vspace{0.3cm}
\caption{\textbf{Zero-shot generalization results on ScanNet\cite{scannet}.} Sup: trained with ground truth, Self-Sup.: trained on image sequences; Pose: uses COLMAP poses at training time.}
\label{tab:scannet}
\end{table}

\textbf{Geometry-Aided Pose Estimation.} 
We now introduce coarse poses estimated by COLMAP. As previously discussed, using such poses seamlessly (j) leads to instability during training, while optimizing the translation component (k) greatly improves the results over (i). 
However, the rotation component estimated by COLMAP is still sub-optimal, preventing GasMono from unleashing its full potential. Indeed, by both optimizing translation and rotation components (l) allows to further reduce the errors.

\textbf{Iterative Self-Distillation.}
To conclude, we verify the impact of ISD. By simply enabling it (m), the accuracy of our architecture starts improving. Moreover, it results effective also when combined with translation optimization (n).  Eventually, enabling ISD to supervise GasMono while optimizing both translation and rotation components (o) yields the overall best results.
To further validate the effectiveness of ISD, we run additional experiments on the KITTI dataset. Tab. \ref{tab:kitti} shows results obtained by plugging distillation strategies -- SD \cite{peng2021excavating} and our ISD -- into Monodepth2 and MonoViT. We can notice how both frameworks are improved, with our method resulting more effective than SD.

\begin{table}[t]
    \centering
    \resizebox{0.47\textwidth}{!}{
    \begin{tabular}{c|cc|cc||cc}
    \hline
    \multirow{2}{*}{Scenes} &
    \multicolumn{2}{c|}{SC-Depthv2~\cite{scdepthv2}} &
    \multicolumn{2}{c||}{\textbf{GasMono (Ours)}} &
    \multicolumn{2}{c}{Monoindoor++$^{*}$ \cite{monoindoor++}}\\
    \cline{2-7}
     & AbsRel$\downarrow$ &  $\delta_1$$\uparrow$ & AbsRel$\downarrow$ &  $\delta_1$$\uparrow$ & AbsRel$\downarrow$ &  $\delta_1 \uparrow$ \\
    \hline
    Chess   & 0.179    & 0.689  & \textbf{0.148} & \textbf{0.791}& 0.157 & 0.750 \\
    Fire    & 0.163    & 0.751  & \textbf{0.131} & \textbf{0.844}& 0.150 & 0.768\\
    Heads   & 0.171    & 0.746 & \textbf{0.151} & \textbf{0.802}& 0.171 & 0.727\\
    Office  & 0.146    & 0.799  & \textbf{0.112} & \textbf{0.878}& 0.130 & 0.837\\
    Pumpkin & 0.120    & 0.841   & 0.136 & 0.852& \textbf{0.102} & \textbf{0.895}\\
    RedKitchen & 0.136 & 0.822   & \textbf{0.130} & \textbf{0.827}& 0.144 & 0.795\\
    Stairs     & \textbf{0.143} & \textbf{0.794}   & 0.151 & 0.782 & 0.155 & 0.753\\
    \hline
    \textbf{Average}     & 0.151 & 0.778 & \textbf{0.137} & \textbf{0.825}& 0.144 & 0.789\\
    \hline
    \end{tabular}
    }
    \vspace{0.3cm}
    
    \caption{\textbf{Zero-shot generalization results on on RGB-D 7-Scenes~\cite{7scene}}. Note that Monoindoor++$^{*}$ \cite{monoindoor++} extracts one image every 30 frames in each video sequence for testset, while we follow SC-Depthv2~\cite{scdepthv2} that extracts the first image every 10 frames. }
    \label{tab:7scenes}
\end{table}

\begin{table}[t]
    \centering
    \resizebox{0.45\textwidth}{!}{
    \begin{tabular}{c|c|c|c|c}
    \hline
    \multirow{2}{*}{Scenes} &
    \multicolumn{4}{c}{\textbf{Fine-tuned}}\\
    \cline{2-5}
    ~ &\multicolumn{2}{c|}{SC-Depthv2~\cite{scdepthv2}} &
    \multicolumn{2}{c}{\textbf{GasMono (Ours)}}  \\
    \cline{2-5}
    & AbsRel$\downarrow$ &  $\delta_1$$\uparrow$& AbsRel$\downarrow$ &  $\delta_1$$\uparrow$ \\
    \hline
    Chess    &0.150 &0.780 & \textbf{0.124} & \textbf{0.867}\\
    Fire     &0.105 &0.918 &\textbf{0.090} &\textbf{0.928}\\
    Heads    &0.143 &0.833 &\textbf{0.111} &\textbf{0.887}\\
    Office   &0.128 &0.855 &\textbf{0.102} &\textbf{0.914}\\
    Pumpkin  &\textbf{0.097} &\textbf{0.922} &0.110 &0.908\\
    RedKitchen  &0.124 &0.853 &\textbf{0.111} &\textbf{0.894}\\
    Stairs      &0.134 &0.823 & \textbf{0.114} & \textbf{0.846}\\
    \hline
    \textbf{Average}    &0.126 &0.854 &\textbf{0.108} &\textbf{0.892}\\
    \hline
    \end{tabular}
    }
    \vspace{0.3cm}
    \caption{\textbf{Fine-tuned results on on RGB-D 7-Scenes~\cite{7scene}}. 
    }
    \label{tab:7scenes_ft}
\end{table}

\begin{figure*}[t]
\begin{center}
 \includegraphics[width=0.98\linewidth]{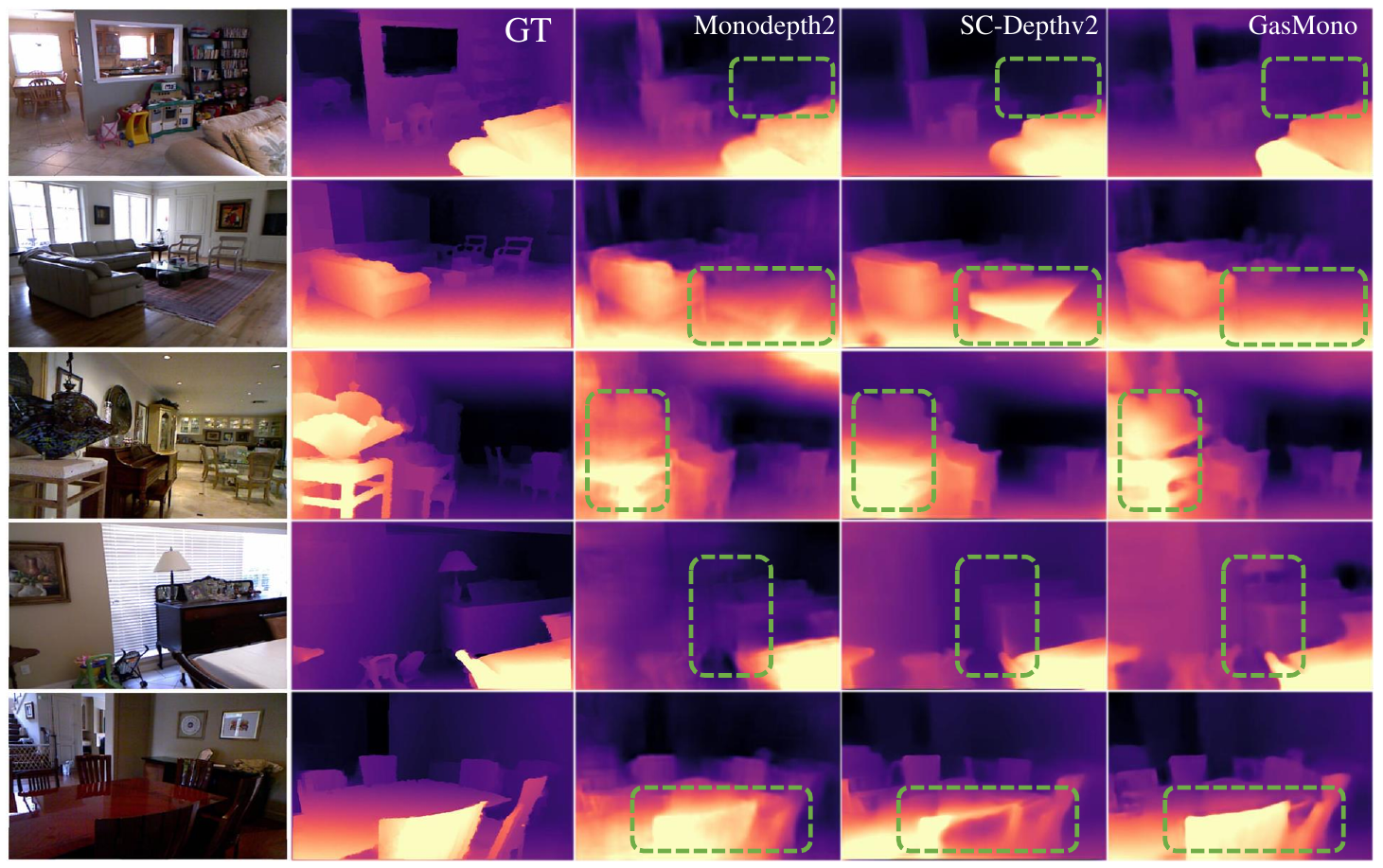}
\end{center}
   \caption{
   \textbf{Qualitative comparison on NYUv2 \cite{nyuv2}.} {GasMono} outperforms the baseline \cite{monodepth2} and the recent SC-Depthv2 \cite{scdepthv2}.}
\label{fig:nyu}
\end{figure*}

\subsection{Comparison with State-of-the-art}
We conclude our experiments by evaluating GasMono with respect to state-of-the-art approaches. 

\textbf{In-Domain Evaluation.} Tab. \ref{tab:nyu} collects results concerning training and testing on the NYUv2 splits. GasMono outperforms existing self-supervised indoor methods \cite{monoindoor++,scdepthv2,wu2022toward} by a large margin,  achieving the best results across all metrics. Especially, compared with our baseline Monodepth2\cite{monodepth2}, our method improves the AbsRel error from
$0.167$ to {0.113} and $\delta_1$ from 75.4\% to {87.1}\%. Moreover, GasMono outperforms the previous top-performing method DistDepth~\cite{wu2022toward}, by {3.9}\% on $\delta_1$, despite this latter is first trained on SimSIN~\cite{wu2022toward} and then fine-tuned on NYUv2, thus exploiting much more training data. 
This allows GasMono for halving the gap between the best self-supervised method and the best supervised one -- DistDepth vs DPT -- on $\delta_1$, reducing it from 6\% to {3}\%.

\textbf{Generalization.} We further test GasMono on ScanNet and 7scenes datasets without retraining. Tabs. \ref{tab:scannet} and \ref{tab:7scenes} collect the outcome of these experiments. On ScanNet, GasMono consistently outperforms any existing self-supervised and supervised framework. On 7scenes, it achieves the best results on 5 out of 7 sets, being the absolute best on average.

For the sake of completeness, we also report the results achieved by fine-tuning GasMono on the 7-scenes dataset. 
Specifically, we follow \cite{scdepthv2} and train for 3 epochs an instance of GasMono for each scene alone, starting from weights pre-trained on NYUv2 dataset \cite{nyuv2}. 
Table \ref{tab:7scenes_ft} shows that our method outperforms SC-Depthv2\cite{scdepthv2}, highlighting that a short fine-tuning process can already improve the performance of our model significantly, with important implications in terms of deployment in real-world applications.

\textbf{Qualitative Results.} Fig. \ref{fig:nyu} shows a qualitative comparison between existing approaches and GasMono. Depth maps predicted by our framework (last column) are more precise, especially in the flat regions 
and fine-grained details. 
More examples in the {{supplementary material}}.

\section{Conclusions}
 This paper proposed GasMono, a geometry-aided self-supervised monocular depth estimation framework for challenging indoor scenes. With the rescale and refine methods proposed in this paper, our method tackles the problem of using coarse poses in self-supervised training, like scale inconsistency between poses, and makes full use of coarse poses from geometry methods. Therefore, our proposal vastly and consistently outperforms any existing approach on the NYUv2 and KITTI datasets. Moreover, experiments on ScanNet and 7Scenes datasets show that GasMono also achieves superior generalization performance.

{\textbf{Acknowledgement:} Work supported in part by the National Key Research and Development Program of China under Grant 2021YFB1714300, National Natural Science Foundation of China under Grant 62233005, the Sino-German Center for Research Promotion under Grant M-0066, the Program of Shanghai Academic Research Leader under Grant 20XD1401300, the Program of Introducing Talents of Discipline to Universities through the 111 Project under Grant B17017 and Shanghai AI Lab.}

{\small
\bibliographystyle{ieee_fullname}
\bibliography{egbib}

\begin{thebibliography}{10}\itemsep=-1pt

\bibitem{bhat2021adabins}
Shariq~Farooq Bhat, Ibraheem Alhashim, and Peter Wonka.
\newblock Adabins: Depth estimation using adaptive bins.
\newblock In {\em CVPR}, 2021.

\bibitem{scdepthv2}
Jia-Wang Bian, Huangying Zhan, Naiyan Wang, Tat-Jin Chin, Chunhua Shen, and Ian Reid.
\newblock Auto-rectify network for unsupervised indoor depth estimation.
\newblock {\em IEEE Transactions on Pattern Analysis and Machine Intelligence (TPAMI)}, 2021.

\bibitem{bian2021unsupervised}
Jia-Wang Bian, Huangying Zhan, Naiyan Wang, Zhichao Li, Le Zhang, Chunhua Shen, Ming-Ming Cheng, and Ian Reid.
\newblock Unsupervised scale-consistent depth learning from video.
\newblock {\em IJCV}, 2021.

\bibitem{scannet}
Angela Dai, Angel~X Chang, Manolis Savva, Maciej Halber, Thomas Funkhouser, and Matthias Nie{\ss}ner.
\newblock Scannet: Richly-annotated 3d reconstructions of indoor scenes.
\newblock In {\em Proceedings of the IEEE conference on computer vision and pattern recognition}, pages 5828--5839, 2017.

\bibitem{imagenet}
Jia Deng, Wei Dong, Richard Socher, Li-Jia Li, Kai Li, and Li Fei-Fei.
\newblock Imagenet: A large-scale hierarchical image database.
\newblock In {\em Conference on Computer Vision and Pattern Recognition (CVPR)}, pages 248--255, Miami, FL, 2009. IEEE.

\bibitem{eigen2015predicting}
David Eigen and Rob Fergus.
\newblock Predicting depth, surface normals and semantic labels with a common multi-scale convolutional architecture.
\newblock In {\em ICCV}, 2015.

\bibitem{eigen2014depth}
David Eigen, Christian Puhrsch, and Rob Fergus.
\newblock Depth map prediction from a single image using a multi-scale deep network.
\newblock {\em NeurIPS}, 2014.

\bibitem{Eigen2014}
David Eigen, Christian Puhrsch, and Rob Fergus.
\newblock Depth map prediction from a single image using a multi-scale deep network.
\newblock {\em Advances in neural information processing systems}, 27, 2014.

\bibitem{fu2018deep}
Huan Fu, Mingming Gong, Chaohui Wang, Kayhan Batmanghelich, and Dacheng Tao.
\newblock Deep ordinal regression network for monocular depth estimation.
\newblock In {\em CVPR}, 2018.

\bibitem{kitti}
Andreas Geiger, Philip Lenz, Christoph Stiller, and Raquel Urtasun.
\newblock Vision meets robotics: The kitti dataset.
\newblock {\em International Journal of Robotics Research (IJRR)}, 2013.

\bibitem{monodepth2}
Clément Godard, Oisin~Mac Aodha, Michael Firman, and Gabriel Brostow.
\newblock Digging into self-supervised monocular depth estimation.
\newblock In {\em International Conference on Computer Vision (ICCV)}, 2019.

\bibitem{godard2017unsupervised}
Cl{\'e}ment Godard, Oisin Mac~Aodha, and Gabriel~J Brostow.
\newblock Unsupervised monocular depth estimation with left-right consistency.
\newblock In {\em CVPR}, 2017.

\bibitem{gonzalez2021plade}
Juan~Luis Gonzalez and Munchurl Kim.
\newblock Plade-net: towards pixel-level accuracy for self-supervised single-view depth estimation with neural positional encoding and distilled matting loss.
\newblock In {\em Proceedings of the IEEE/CVF Conference on Computer Vision and Pattern Recognition}, pages 6851--6860, 2021.

\bibitem{gonzalezbello2020forget}
Juan~Luis GonzalezBello and Munchurl Kim.
\newblock Forget about the lidar: Self-supervised depth estimators with med probability volumes.
\newblock {\em Advances in Neural Information Processing Systems}, 33:12626--12637, 2020.

\bibitem{guizilini20203d}
Vitor Guizilini, Rares Ambrus, Sudeep Pillai, Allan Raventos, and Adrien Gaidon.
\newblock {3D} packing for self-supervised monocular depth estimation.
\newblock In {\em CVPR}, 2020.

\bibitem{brnet}
Wencheng Han, Junbo Yin, Xiaogang Jin, Xiangdong Dai, and Jianbing Shen.
\newblock Brnet: Exploring comprehensive features for monocular depth estimation.
\newblock In {\em European Conference on Computer Vision}, pages 586--602. Springer, 2022.

\bibitem{hartley2003multiple}
Richard Hartley and Andrew Zisserman.
\newblock {\em Multiple view geometry in computer vision}.
\newblock Cambridge university press, 2003.

\bibitem{resnet}
Kaiming He, Xiangyu Zhang, Shaoqing Ren, and Jian Sun.
\newblock Deep residual learning for image recognition.
\newblock In {\em Conference on Computer Vision and Pattern Recognition (CVPR)}, pages 770--778, Las Vegas, Nevada, 2016. IEEE.

\bibitem{ji2021refine}
Mingi Ji, Seungjae Shin, Seunghyun Hwang, Gibeom Park, and Il-Chul Moon.
\newblock Refine myself by teaching myself: Feature refinement via self-knowledge distillation.
\newblock In {\em Proceedings of the IEEE/CVF Conference on Computer Vision and Pattern Recognition}, pages 10664--10673, 2021.

\bibitem{ji2021monoindoor}
Pan Ji, Runze Li, Bir Bhanu, and Yi Xu.
\newblock Monoindoor: Towards good practice of self-supervised monocular depth estimation for indoor environments.
\newblock In {\em ICCV}, 2021.

\bibitem{klingner2020self}
Marvin Klingner, Jan-Aike Term{\"o}hlen, Jonas Mikolajczyk, and Tim Fingscheidt.
\newblock Self-supervised monocular depth estimation: Solving the dynamic object problem by semantic guidance.
\newblock In {\em ECCV}, 2020.

\bibitem{Klodt_2018_ECCV}
Maria Klodt and Andrea Vedaldi.
\newblock Supervising the new with the old: learning sfm from sfm.
\newblock In {\em Proceedings of the European Conference on Computer Vision (ECCV)}, September 2018.

\bibitem{laina2016deeper}
Iro Laina, Christian Rupprecht, Vasileios Belagiannis, Federico Tombari, and Nassir Navab.
\newblock Deeper depth prediction with fully convolutional residual networks.
\newblock In {\em 3DV}, 2016.

\bibitem{mpvit}
Youngwan Lee, Jonghee Kim, Jeffrey Willette, and Sung~Ju Hwang.
\newblock Mpvit: Multi-path vision transformer for dense prediction.
\newblock In {\em Proceedings of the IEEE/CVF Conference on Computer Vision and Pattern Recognition}, pages 7287--7296, 2022.

\bibitem{li2021structdepth}
Boying Li, Yuan Huang, Zeyu Liu, Danping Zou, and Wenxian Yu.
\newblock Structdepth: Leveraging the structural regularities for self-supervised indoor depth estimation.
\newblock In {\em ICCV}, 2021.

\bibitem{li2017two}
Jun Li, Reinhard Klein, and Angela Yao.
\newblock A two-streamed network for estimating fine-scaled depth maps from single rgb images.
\newblock In {\em CVPR}, 2017.

\bibitem{monoindoor++}
Runze Li, Pan Ji, Yi Xu, and Bir Bhanu.
\newblock Monoindoor++: Towards better practice of self-supervised monocular depth estimation for indoor environments.
\newblock {\em IEEE Transactions on Circuits and Systems for Video Technology}, 2022.

\bibitem{liu2022accurate}
Chenxin Liu, Jiahu Qin, Shuai Wang, Lei Yu, and Yaonan Wang.
\newblock Accurate rgb-d slam in dynamic environments based on dynamic visual feature removal.
\newblock {\em Science China Information Sciences}, 65(10):202206, 2022.

\bibitem{adamw}
Ilya Loshchilov and Frank Hutter.
\newblock Decoupled weight decay regularization.
\newblock {\em arXiv preprint arXiv:1711.05101}, 2017.

\bibitem{hrdepth}
Xiaoyang Lyu, Liang Liu, Mengmeng Wang, Xin Kong, Lina Liu, Yong Liu, Xinxin Chen, and Yi Yuan.
\newblock Hr-depth: High resolution self-supervised monocular depth estimation.
\newblock {\em In AAAI Conference on Artificial Intelligence (AAAI)}, 2021.

\bibitem{orbslam}
Raul Mur-Artal, Jose Maria~Martinez Montiel, and Juan~D Tardos.
\newblock Orb-slam: a versatile and accurate monocular slam system.
\newblock {\em IEEE transactions on robotics}, 31(5):1147--1163, 2015.

\bibitem{nyuv2}
Pushmeet~Kohli Nathan~Silberman, Derek~Hoiem and Rob Fergus.
\newblock Indoor segmentation and support inference from rgbd images.
\newblock In {\em ECCV}, 2012.

\bibitem{peng2021excavating}
Rui Peng, Ronggang Wang, Yawen Lai, Luyang Tang, and Yangang Cai.
\newblock Excavating the potential capacity of self-supervised monocular depth estimation.
\newblock In {\em Proceedings of the IEEE/CVF International Conference on Computer Vision}, pages 15560--15569, 2021.

\bibitem{poggi2020uncertainty}
Matteo Poggi, Filippo Aleotti, Fabio Tosi, and Stefano Mattoccia.
\newblock On the uncertainty of self-supervised monocular depth estimation.
\newblock In {\em CVPR}, 2020.

\bibitem{poggi2021synergies}
Matteo Poggi, Fabio Tosi, Konstantinos Batsos, Philippos Mordohai, and Stefano Mattoccia.
\newblock On the synergies between machine learning and binocular stereo for depth estimation from images: a survey.
\newblock {\em IEEE Transactions on Pattern Analysis and Machine Intelligence}, 44(9):5314--5334, 2021.

\bibitem{3net18}
Matteo Poggi, Fabio Tosi, and Stefano Mattoccia.
\newblock Learning monocular depth estimation with unsupervised trinocular assumptions.
\newblock In {\em 6th International Conference on 3D Vision (3DV)}, 2018.

\bibitem{qiao2022improving}
Hong Qiao, Shanlin Zhong, Ziyu Chen, and Hongze Wang.
\newblock Improving performance of robots using human-inspired approaches: a survey.
\newblock {\em Science China Information Sciences}, 65(12):221201, 2022.

\bibitem{Ranftl2021}
Ren\'{e} Ranftl, Alexey Bochkovskiy, and Vladlen Koltun.
\newblock Vision transformers for dense prediction.
\newblock {\em ICCV}, 2021.

\bibitem{Ranftl2020}
Ren\'{e} Ranftl, Katrin Lasinger, David Hafner, Konrad Schindler, and Vladlen Koltun.
\newblock Towards robust monocular depth estimation: Mixing datasets for zero-shot cross-dataset transfer.
\newblock {\em TPAMI}, 2020.

\bibitem{saxena2008make3d}
Ashutosh Saxena, Min Sun, and Andrew~Y Ng.
\newblock {Make3D}: Learning {3D} scene structure from a single still image.
\newblock {\em IEEE transactions on pattern analysis and machine intelligence}, 31(5):824--840, 2008.

\bibitem{colmap}
Johannes~Lutz Sch\"{o}nberger and Jan-Michael Frahm.
\newblock Structure-from-motion revisited.
\newblock In {\em Conference on Computer Vision and Pattern Recognition (CVPR)}, 2016.

\bibitem{7scene}
Jamie Shotton, Ben Glocker, Christopher Zach, Shahram Izadi, Antonio Criminisi, and Andrew Fitzgibbon.
\newblock Scene coordinate regression forests for camera relocalization in rgb-d images.
\newblock In {\em Proceedings of the IEEE conference on computer vision and pattern recognition}, pages 2930--2937, 2013.

\bibitem{sun2021unsupervised}
Qiyu Sun, Yang Tang, Chongzhen Zhang, Chaoqiang Zhao, Feng Qian, and J{\"u}rgen Kurths.
\newblock Unsupervised estimation of monocular depth and vo in dynamic environments via hybrid masks.
\newblock {\em IEEE Transactions on Neural Networks and Learning Systems}, 2021.

\bibitem{tang2022perception}
Yang Tang, Chaoqiang Zhao, Jianrui Wang, Chongzhen Zhang, Qiyu Sun, Wei~Xing Zheng, Wenli Du, Feng Qian, and J{\"u}rgen Kurths.
\newblock Perception and navigation in autonomous systems in the era of learning: A survey.
\newblock {\em IEEE Transactions on Neural Networks and Learning Systems}, 2022.

\bibitem{raft}
Zachary Teed and Jia Deng.
\newblock Raft: Recurrent all-pairs field transforms for optical flow.
\newblock In {\em European conference on computer vision}, pages 402--419. Springer, 2020.

\bibitem{Tosi_2019_CVPR}
Fabio Tosi, Filippo Aleotti, Matteo Poggi, and Stefano Mattoccia.
\newblock Learning monocular depth estimation infusing traditional stereo knowledge.
\newblock In {\em The IEEE Conference on Computer Vision and Pattern Recognition (CVPR)}, June 2019.

\bibitem{tosi2020distilled}
Fabio Tosi, Filippo Aleotti, Pierluigi~Zama Ramirez, Matteo Poggi, Samuele Salti, Luigi Di~Stefano, and Stefano Mattoccia.
\newblock Distilled semantics for comprehensive scene understanding from videos.
\newblock In {\em Proceedings of the IEEE Conference on Computer Vision and Pattern Recognition}, 2020.

\bibitem{watson2019self}
Jamie Watson, Michael Firman, Gabriel~J Brostow, and Daniyar Turmukhambetov.
\newblock Self-supervised monocular depth hints.
\newblock In {\em ICCV}, 2019.

\bibitem{wu2022toward}
Cho-Ying Wu, Jialiang Wang, Michael Hall, Ulrich Neumann, and Shuochen Su.
\newblock Toward practical monocular indoor depth estimation.
\newblock In {\em Proceedings of the IEEE/CVF Conference on Computer Vision and Pattern Recognition}, pages 3814--3824, 2022.

\bibitem{cadepth}
Jiaxing Yan, Hong Zhao, Penghui Bu, and YuSheng Jin.
\newblock Channel-wise attention-based network for self-supervised monocular depth estimation.
\newblock In {\em 2021 International Conference on 3D Vision (3DV)}, pages 464--473. IEEE, 2021.

\bibitem{drivingstereo}
Guorun Yang, Xiao Song, Chaoqin Huang, Zhidong Deng, Jianping Shi, and Bolei Zhou.
\newblock Drivingstereo: A large-scale dataset for stereo matching in autonomous driving scenarios.
\newblock In {\em IEEE Conference on Computer Vision and Pattern Recognition (CVPR)}, 2019.

\bibitem{Yang_CVPR_2018}
Zhenheng Yang, Peng Wang, Wang Yang, Wei Xu, and Nevatia Ram.
\newblock Lego: Learning edge with geometry all at once by watching videos.
\newblock In {\em The IEEE Conference on Computer Vision and Pattern Recognition (CVPR)}, 2018.

\bibitem{vnl}
Wei Yin, Yifan Liu, Chunhua Shen, and Youliang Yan.
\newblock Enforcing geometric constraints of virtual normal for depth prediction.
\newblock In {\em Proceedings of the IEEE/CVF International Conference on Computer Vision}, pages 5684--5693, 2019.

\bibitem{yin2018geonet}
Zhichao Yin and Jianping Shi.
\newblock Geonet: Unsupervised learning of dense depth, optical flow and camera pose.
\newblock In {\em Proceedings of the IEEE conference on computer vision and pattern recognition}, pages 1983--1992, 2018.

\bibitem{yu2020p}
Zehao Yu, Lei Jin, and Shenghua Gao.
\newblock P$^2$net: Patch-match and plane-regularization for unsupervised indoor depth estimation.
\newblock In {\em ECCV}, 2020.

\bibitem{zhang2023qa}
Hongwei Zhang, Wenbo Xiong, Shuai Lu, Mengdan Chen, and Le Yao.
\newblock Qa-ustnet: Yarn-dyed fabric defect detection via u-shaped swin transformer network based on quadtree attention.
\newblock {\em Textile Research Journal}, page 00405175231158134, 2023.

\bibitem{zhao2020monocular}
Chaoqiang Zhao, Qiyu Sun, Chongzhen Zhang, Yang Tang, and Feng Qian.
\newblock Monocular depth estimation based on deep learning: An overview.
\newblock {\em Science China Technological Sciences}, 63(9):1612--1627, 2020.

\bibitem{zhao2020masked}
Chaoqiang Zhao, Gary~G Yen, Qiyu Sun, Chongzhen Zhang, and Yang Tang.
\newblock Masked gan for unsupervised depth and pose prediction with scale consistency.
\newblock {\em IEEE Transactions on Neural Networks and Learning Systems}, 32(12):5392--5403, 2020.

\bibitem{monovit}
Chaoqiang Zhao, Youmin Zhang, Matteo Poggi, Fabio Tosi, Xianda Guo, Zheng Zhu, Guan Huang, Yang Tang, and Stefano Mattoccia.
\newblock Monovit: Self-supervised monocular depth estimation with a vision transformer.
\newblock In {\em 2022 International Conference on 3D Vision (3DV)}. IEEE, 2022.

\bibitem{zhao2020towards}
Wang Zhao, Shaohui Liu, Yezhi Shu, and Yong-Jin Liu.
\newblock Towards better generalization: Joint depth-pose learning without posenet.
\newblock In {\em CVPR}, 2020.

\bibitem{diffnet}
Hang Zhou, David Greenwood, and Sarah Taylor.
\newblock Self-supervised monocular depth estimation with internal feature fusion.
\newblock In {\em British Machine Vision Conference (BMVC)}, 2021.

\bibitem{zhou2019moving}
Junsheng Zhou, Yuwang Wang, Kaihuai Qin, and Wenjun Zeng.
\newblock Moving indoor: Unsupervised video depth learning in challenging environments.
\newblock In {\em CVPR}, 2019.

\bibitem{devnet}
Kaichen Zhou, Lanqing Hong, Changhao Chen, Hang Xu, Chaoqiang Ye, Qingyong Hu, and Zhenguo Li.
\newblock Devnet: Self-supervised monocular depth learning via density volume construction.
\newblock {\em European Conference on Computer Vision}, 2022.

\bibitem{zhou2017unsupervised}
Tinghui Zhou, Matthew Brown, Noah Snavely, and David~G Lowe.
\newblock Unsupervised learning of depth and ego-motion from video.
\newblock In {\em Proceedings of the IEEE conference on computer vision and pattern recognition}, pages 1851--1858, 2017.

\bibitem{zhou2019continuity}
Yi Zhou, Connelly Barnes, Jingwan Lu, Jimei Yang, and Hao Li.
\newblock On the continuity of rotation representations in neural networks.
\newblock In {\em Proceedings of the IEEE/CVF Conference on Computer Vision and Pattern Recognition}, pages 5745--5753, 2019.

\end{thebibliography}
}

\newpage\phantom{Supplementary}
\multido{\i=1+1}{6}{


\section*{Supplementary material}

# GasMono: Geometry-Aided Self-Supervised Monocular Depth Estimation for Indoor Scenes

[1]Chaoqiang Zhao    [2]Matteo Poggi    [2]Fabio Tosi
[1]Lei Zhou    [1]Qiyu Sun    [1,*]Yang Tang    [2]Stefano Mattoccia

[1]East China University of Science and Technology    [2]University of Bologna

This document provides additional details for ICCV paper "GasMono: Geometry-Aided Self-Supervised monocular depth estimation for indoor scenes".

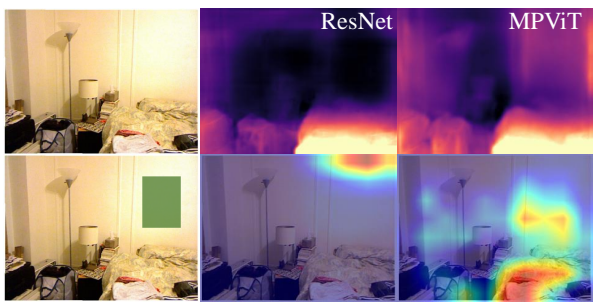

Figure 1. **Attention maps by different encoders.** On top: RGB image and depth map by models using CNN-based (ResNet) and transformer-based (MPViT) encoders. At bottom: untextured region highlighted on RGB image and attention maps by ResNet and MPViT based models.

## 1. Ablation studies

For the sake of space, we complement here the ablation studies reported in the main paper.

**Encoder:** Benefiting from the global feature extraction of the transformer, the model can infer the depth of low-texture regions more accurately text to global attention. Fig. 1 visually shows an intuition of this mechanism: on left, we show an RGB image with a large, untextured wall, highlighted in green on the second row. A model with a ResNet encoder cannot estimate an accurate depth map (second column), as the attention of the last features is mainly focused on the low-texture area on top of the image. Conversely, the output of a model using MPVit as a backbone results coherent, thanks to the global guidance it can recover from the bottom regions of the image showing richer texture. Furthermore, we report an ablation study in Tab. 1 comparing the results yielded by MPViT variants [5] (tiny, xsmall, small, base). We identify a good trade-off between accuracy and complexity in "MPViT-small", which we use as

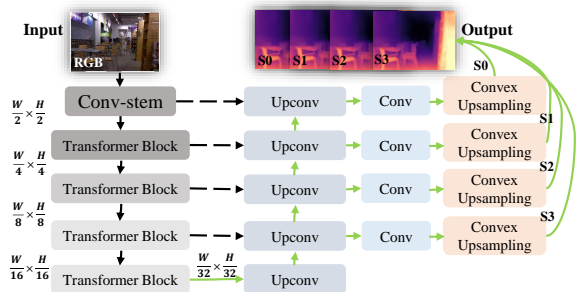

Figure 2. **Architecture of Depth Network.** The network takes a single RGB image as input and outputs four full-resolution disparity maps [S0, S1, S2, S3].

the encoder backbone for all our experiments in the main paper.

**Low texture regions:** To further verify the effectiveness of different encoders on low-texture regions, we evaluate the depth estimation performance of models on low-texture regions. We use image gradient information to select and locate low-texture regions, as shown in Fig.4. As shown in Table 1 (V-X), with the global modeling ability of the transformer, the MPViT encoder shows outstanding performance on low-texture regions than CNN encoders.

**Decoder:** For the depth decoder, we follow the architecture design proposed by Monodepth2 [3], yet we implement convex upsampling as proposed by RAFT [8], to output four full resolution disparity maps, i.e. depth maps [S0, S1, S2, S3] in Fig. 2. We compare the accuracy yielded by the standard decoders proposed by Monodepth2 [3] and ours in Table 1, rows (E-H). We can notice that our decoder allows for improving the results by both Monodepth2 itself and our network architecture.

**ISD:** As shown in Table 1 (K-R), with the help of the iterative self-distillation method we proposed in this paper, depth maps are accurate at any scale. Note that the depth map **S3** is as accurate as the depth map **S0**, which means that the final features extracted by the encoder include the

| Details | Params | lower is better | | | | higher is better | | |
|---|---|---|---|---|---|---|---|---|
| | | Abs Rel↓ | Sq Rel↓ | RMSE↓ | RMSE log ↓ | $\delta_1$ ↑ | $\delta_2$ ↑ | $\delta_3$ ↑ |
| | | **Different ResNet and MPViT [5] variants** | | | | | | |
| (A) ResNet–18 | 11.6M | 0.141 | 0.107 | 0.533 | 0.183 | 0.812 | 0.957 | 0.989 |
| (B) ResNet–34 | 21.8M | 0.133 | 0.100 | 0.517 | 0.174 | 0.828 | 0.963 | 0.991 |
| (C) MPViT-**tiny** | 5.8M | 0.122 | 0.087 | 0.478 | 0.162 | 0.854 | 0.969 | 0.992 |
| (D) MPViT-**xsmall** | 10.5M | 0.117 | 0.085 | 0.472 | 0.157 | 0.863 | 0.972 | 0.992 |
| (E) MPViT-**small** | 22.8M | **0.113** | **0.083** | **0.459** | **0.153** | 0.871 | **0.973** | 0.992 |
| (F) MPViT-**base** | 74.8M | **0.113** | 0.085 | 0.466 | **0.153** | **0.872** | **0.973** | **0.993** |

| Details | Decoder | Different decoders | | | | | | |
|---|---|---|---|---|---|---|---|---|
| (G) Monodepth2 [3] | Monodepth2 [3] | 0.167 | 0.137 | 0.603 | 0.208 | 0.754 | 0.944 | 0.985 |
| (H) Monodepth2 [3] | ours | 0.161 | 0.130 | 0.586 | 0.202 | 0.767 | 0.949 | 0.988 |
| (I) Ours | Monodepth2 [3] | 0.114 | 0.086 | 0.465 | 0.154 | 0.869 | 0.972 | 0.992 |
| (J) Ours | ours | 0.113 | 0.083 | 0.459 | 0.153 | 0.871 | 0.973 | 0.992 |

| Details | Scale | Disparity maps from different scale | | | | | | |
|---|---|---|---|---|---|---|---|---|
| (K) Ours | S0 | 0.113 | 0.083 | 0.459 | 0.153 | 0.871 | 0.973 | 0.992 |
| (L) Ours(w/o **ISD**) | S0 | 0.114 | 0.085 | 0.469 | 0.156 | 0.867 | 0.972 | 0.992 |
| (M) Ours | S1 | 0.114 | 0.082 | 0.459 | 0.153 | 0.870 | 0.973 | 0.993 |
| (N) Ours( w/o **ISD**) | S1 | 0.139 | 0.099 | 0.522 | 0.177 | 0.822 | 0.968 | 0.994 |
| (O) Ours | S2 | 0.114 | 0.082 | 0.458 | 0.153 | 0.869 | 0.973 | 0.993 |
| (P) Ours( w/o **ISD**) | S2 | 0.175 | 0.299 | 0.918 | 0.231 | 0.791 | 0.947 | 0.980 |
| (Q) Ours | S3 | 0.114 | 0.082 | 0.460 | 0.153 | 0.868 | 0.973 | 0.993 |
| (R) Ours( w/o **ISD**) | S3 | 0.203 | 0.534 | 1.122 | 0.272 | 0.787 | 0.943 | 0.972 |

| Details | n \| training time | Results under different number of iterations | | | | | | |
|---|---|---|---|---|---|---|---|---|
| (S) Ours | **1** \| 26.2h | 0.115 | 0.084 | 0.467 | 0.155 | 0.867 | 0.973 | 0.992 |
| (T) Ours | **2** \| 44.3 h | **0.113** | 0.083 | **0.459** | **0.153** | 0.871 | **0.973** | **0.992** |
| (U) Ours | **3** \| 64.1h | **0.112** | **0.082** | 0.460 | **0.152** | **0.873** | 0.973 | **0.992** |

| Details | test regions | Results of different encoders on low texture regions | | | | | | |
|---|---|---|---|---|---|---|---|---|
| (V) R18 (GasMono training framework) | | 0.140 | 0.106 | 0.527 | 0.181 | 0.816 | 0.958 | 0.990 |
| (W) R34 (GasMono training framework) | Low Texture | 0.131 | 0.099 | 0.512 | 0.172 | 0.834 | 0.963 | 0.991 |
| (X) MPViT-**small** (GasMono training framework) | | 0.110 | 0.082 | 0.453 | 0.149 | 0.877 | 0.974 | 0.993 |

Table 1. **Ablation Studies on NYUv2.** We test the depth encoder based on different MPViT variants and different depth decoders on NYUv2 datasets. The effects of **ISD** on the accuracy of each scale depth map are also evaluated. We also test the performance of different encoders on low-texture regions.

| | Depth Encoder | Pose Estimation | lower is better | | | | higher is better | | |
|---|---|---|---|---|---|---|---|---|---|
| | | | Abs Rel↓ | Sq Rel↓ | RMSE↓ | RMSE log ↓ | $\delta_1$ ↑ | $\delta_2$ ↑ | $\delta_3$ ↑ |
| (i1) | ResNet–18 | Monoindoor++ [6] | 0.149 | 0.116 | 0.551 | 0.190 | 0.800 | 0.953 | 0.988 |
| (i2) | ResNet–18 | GasMono(ours) | 0.141 | 0.107 | 0.533 | 0.183 | 0.812 | 0.957 | 0.989 |
| (ii1) | MPViT-**small** | Monoindoor++ [6] | 0.124 | 0.092 | 0.488 | 0.163 | 0.851 | 0.970 | 0.992 |
| (ii2) | MPViT-**small** | GasMono(ours) | 0.113 | 0.083 | 0.459 | 0.153 | 0.871 | 0.973 | 0.992 |

Table 2. **Ablation Studies of different pose estimation methods on NYUv2 dataset**. To compare with the SoTA method – monoindoor++, we replace the pose estimation method proposed in GasMono with that proposed in monoindoor++. Except for the pose estimation part, the settings of the training process are all the same.

key information for depth inferring, and with the help of skip connection, the depth is optimized as information increases. Besides, we also do an ablation study on the number of iterations **n**, as shown in Table 1 (S-U). When **n** = 1, the self-distillation method cannot improve the accuracy of the depth labels already obtained by the framework itself, having a negative impact on the performance. When using **n** > 1, ISD allows for getting better pseudo labels, thereby yielding a much better depth model after training. In the main paper, we select **n** = 2, since further increases do not yield substantial improvement, yet increasing notably the total training time.

## 2. Coarse Pose analysis

We now further inquire about the use of coarse poses obtained through COLMAP.

**COLMAP for coarse poses:** COLMAP is executed off-line, so it does not affect training/inference time directly. It only delays the start of the training (on NYUv2, a single sequence takes ∼ 10 min, with sequences being processed in parallel), a delay we can tolerate given the SoTA accuracy it yields. Since some sequences in NYUv2 are very short, and some scenes (like the bathroom) expose many low-texture regions, COLMAP may fail on some of these sequences, which are excluded from the training set training.

One of the most important and valuable points in this paper is: the scale ambiguity observed across the hundred se-

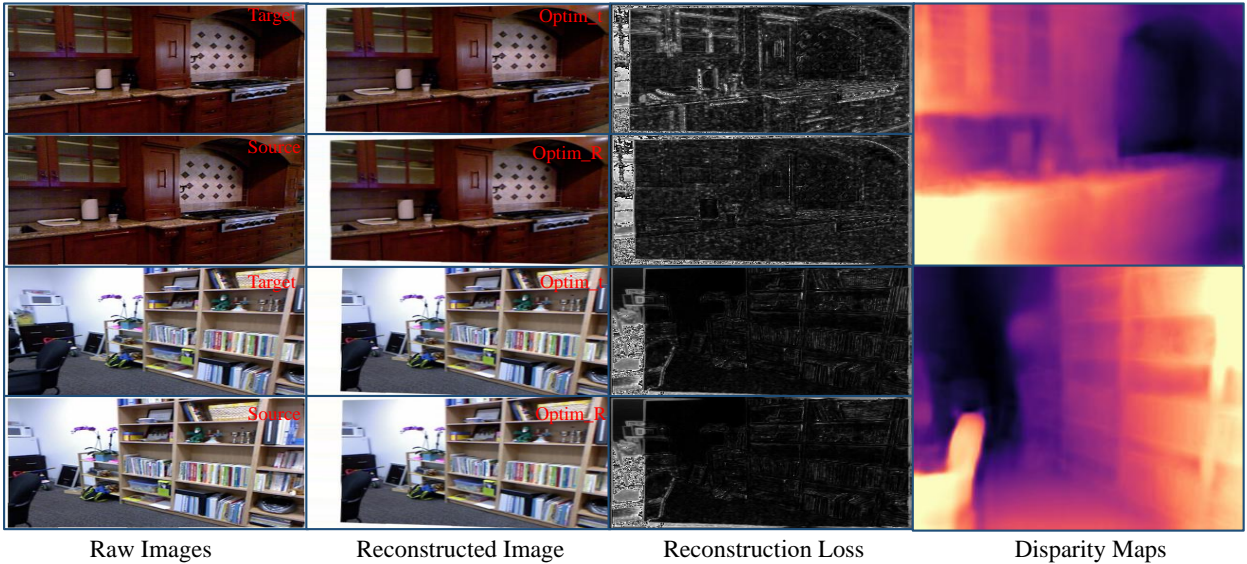

Figure 3. **Visualization of the coarse rotation optimization.** The two samples are based on inaccurate and accurate coarse rotations, and both the rotations between the source and target images are large, resulting in large unreconstructable regions. With the accurate coarse rotation, the coarse pose with translation optimization ($Optim\_t$) can get a satisfactory reconstruction, see 2nd sample. While the coarse rotation is inaccurate, we need to improve the rotation further, otherwise, it will introduce noise into training, see the reconstruction loss of 1st sample. Besides, as shown in the reconstruction loss, the rotation from the coarse pose can eliminate most rotation components, which makes it possible to directly estimate the residual rotation using a pose network.

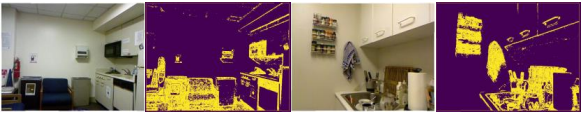

Figure 4. Low texture regions for depth evaluation.

quences used for training makes COLMAP use not as simple/effective as expected, avoiding network convergence – see Tab.1 ($a$-$g$) in the paper. Indeed, using COLMAP poses alone is, counter-intuitively, harmful to indoor SMDE.

**Translation rescaling and refinement:** By its nature, our AlignNet learns to regress a scale value that is meaningful for training images only – i.e., it aligns the scale of the translation component of COLMAP poses on the training set, to allow for DepthNet convergence (Tab. 1). As such, it is a training optimization tool only, and it would be ineffective on test images. Therefore, considering the tightly coupled relationship between the depth estimation part and the pose estimation part, we use the performance of the depth network to verify the effectiveness of pose optimization – see Tab.1 ($a$-$g$) in the paper. After **translation rescaling and refinement**, our GasMono has achieved SoTA performance – see Tab.1 ($n$) in the paper.

**Rotation optimization:** The above part only focuses on translation optimization, though the rotation of coarse poses can also be inaccurate and noisy, so rotation optimization is proposed. To further show the effect of rotation optimiza-

tion within the training, in Fig. 3, we report two samples with inaccurate coarse rotation (top sample) and accurate coarse rotation (bottom sample) respectively. For both, we compute the reconstruction losses based on "Optim\_t" and "Optim\_R" and show them in column 3, based on estimated depth (column 4). For the first sample, because of the inaccurate coarse rotation, optimizing only for translation ("Optim\_t", row 1) cannot compensate for the wrong rotation and thus yield a high reprojection error. After refining the rotation, the reconstruction based on "Optim\_R" (row 2) results in a much lower photometric error. On the contrary, in case of accurate coarse pose, shown as the second sample in the figure, the reconstruction based on "Optim\_t" can already achieve a reasonable reconstructed image.

**Comparison with different pose estimation strategies:** To further prove the effectiveness of the pose estimation part in our GasMono, we compare it with the SoTA self-supervised indoor framework – monoindoor++ [6]. Since the SoTA Monoindoor series [4, 6] do not open source their code, we cannot construct experiments on the Monoindoor framework. Therefore, based on our GasMono, we replace the pose estimation method with the one proposed in monoindoor++, which is purely the learning-based pose estimation method. Except for the pose estimation part, the settings of the training process and loss functions are all the same. Experimental results in Tab. 2 show that our proposed method yields more accurate results, which means that the proposed geometric-assisted

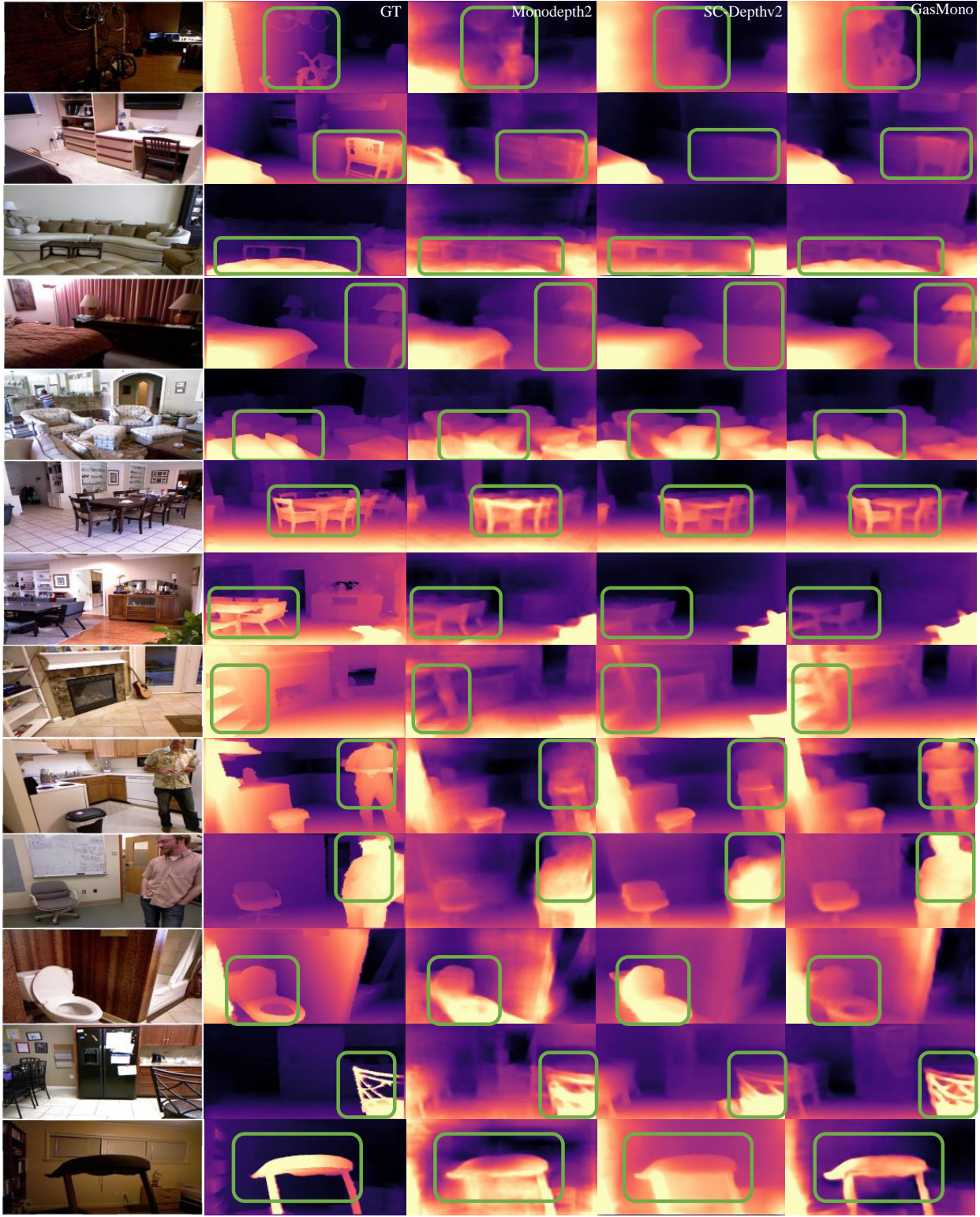

Figure 5. **Visualization on NYUv2.** Our **GasMono** achieved finer-grained depth estimation than the baseline Monodepth2 [3] and recent work SC-Depthv2 [1].

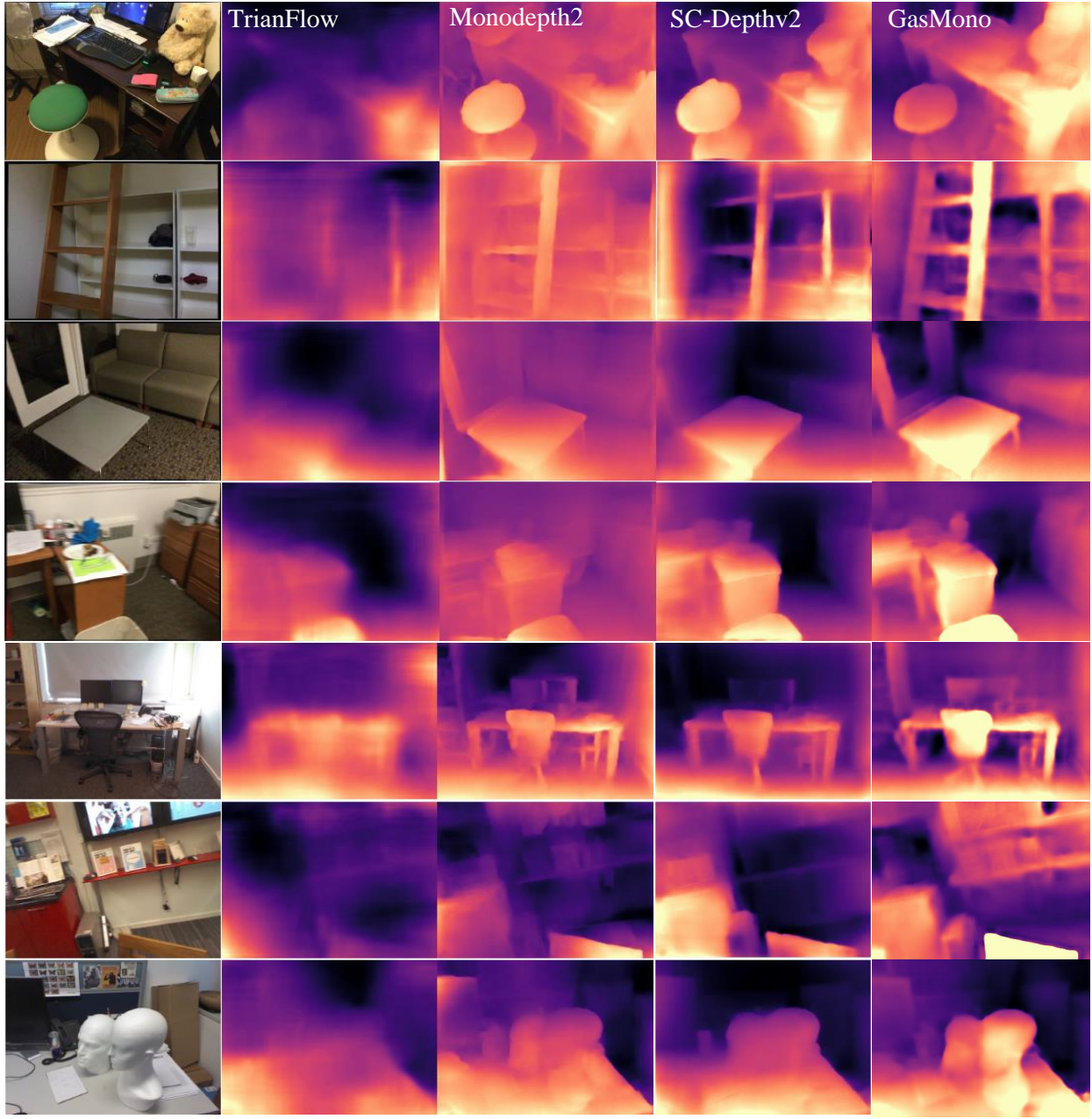

Figure 6. **Visualization on ScanNet [2] and 7scenes [7].** Generalization comparison with the Trianflow [9],Monodepth2 [3] and SC-Depthv2 [1]. GasMono shows more accurate and finer-grained depth estimation on new scenes.

training framework as well as the proposed coarse pose optimization method are more practical, effective, and beneficial for the indoor self-supervised monocular depth estimation task.

## 3. Qualitative results

We report additional qualitative results to show a comparison with existing self-supervised frameworks.

**Comparison with other methods – in-domain:** We show some qualitative results in Fig. 5, comparing some existing methods with our GasMono framework. Ours outperforms the baseline Monodepth2 [3] and a more recent work, i.e. SC-Depthv2 [1], on fine-grained depth estimation under different conditions. GasMono shows outstanding performance when estimating depth at object details, such as the bicycle (1st line), chair (2nd and 6-7th lines),

desk lamp (4th line), and so on. Besides, for large-scale objects, like the sofa (3rd line), chair (12th), and treadmill (13th), and dynamic objects, like people (9-10th lines), our model shows very accurate results as well.

**Comparison with other methods – generalization:** We also show more qualitative examples for generalization to unseen datasets in Fig. 6. Compared with recent works [1, 3, 9], GasMono shows higher-quality results in estimating the depth for unseen objects.

## 4. Limitations

Since our training framework builds on self-supervised losses, GasMono still cannot perform well in regions where reprojection constraints are violated, e.g. specular/transparent surfaces or surfaces with reflected light. As shown in Fig. 7, since those regions do not satisfy the photometric consistency assumption, GasMono fails on highly-specular or transparent surfaces.

Moreover, running COLMAP to obtain the coarse poses – required to start the training – is time-consuming, in particular on datasets counting hundred of sequences made of thousand frames each.

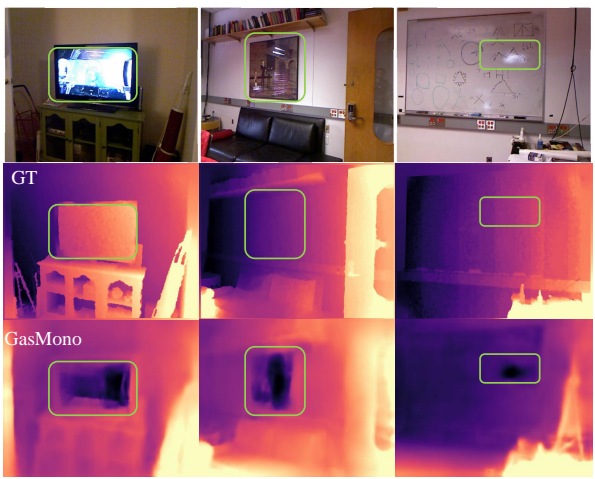

Figure 7. **Limitations of GasMono.** Since GasMono is trained in a self-supervised manner, it cannot estimate a reasonable depth for those regions with photometric changes.

## References

[1] Jia-Wang Bian, Huangying Zhan, Naiyan Wang, Tat-Jin Chin, Chunhua Shen, and Ian Reid. Auto-rectify network for unsupervised indoor depth estimation. *IEEE Transactions on Pattern Analysis and Machine Intelligence (TPAMI)*, 2021. 4, 5, 6

[2] Angela Dai, Angel X Chang, Manolis Savva, Maciej Halber, Thomas Funkhouser, and Matthias Nießner. Scannet: Richly-annotated 3d reconstructions of indoor scenes. In *Proceedings of the IEEE conference on computer vision and pattern recognition*, pages 5828–5839, 2017. 5

[3] Clément Godard, Oisin Mac Aodha, Michael Firman, and Gabriel Brostow. Digging into self-supervised monocular depth estimation. In *International Conference on Computer Vision (ICCV)*, 2019. 1, 2, 4, 5, 6

[4] Pan Ji, Runze Li, Bir Bhanu, and Yi Xu. Monoindoor: Towards good practice of self-supervised monocular depth estimation for indoor environments. In *ICCV*, 2021. 3

[5] Youngwan Lee, Jonghee Kim, Jeffrey Willette, and Sung Ju Hwang. Mpvit: Multi-path vision transformer for dense prediction. In *Proceedings of the IEEE/CVF Conference on Computer Vision and Pattern Recognition*, pages 7287–7296, 2022. 1, 2

[6] Runze Li, Pan Ji, Yi Xu, and Bir Bhanu. Monoindoor++: Towards better practice of self-supervised monocular depth estimation for indoor environments. *IEEE Transactions on Circuits and Systems for Video Technology*, 2022. 2, 3

[7] Jamie Shotton, Ben Glocker, Christopher Zach, Shahram Izadi, Antonio Criminisi, and Andrew Fitzgibbon. Scene coordinate regression forests for camera relocalization in rgb-d images. In *Proceedings of the IEEE conference on computer vision and pattern recognition*, pages 2930–2937, 2013. 5

[8] Zachary Teed and Jia Deng. Raft: Recurrent all-pairs field transforms for optical flow. In *European conference on computer vision*, pages 402–419. Springer, 2020. 1

[9] Wang Zhao, Shaohui Liu, Yezhi Shu, and Yong-Jin Liu. Towards better generalization: Joint depth-pose learning without posenet. In *CVPR*, 2020. 5, 6


}

\end{document}